\documentclass[preprint,review]{elsarticle}
\usepackage{color,xcolor}

\usepackage{epsfig}
\usepackage{graphicx}
\usepackage{amsmath}
\usepackage{amssymb}
\usepackage{float}
\usepackage{booktabs}
\usepackage{algorithm}
\usepackage{algorithmic}
\usepackage{booktabs}
\usepackage{times}

\usepackage{graphicx}
\usepackage{amsmath}
\usepackage{amssymb}
\usepackage{amsthm}
\usepackage{url}
\usepackage{array}
\usepackage{soul}
\newcolumntype{P}[1]{>{\centering\arraybackslash}p{#1}}
\newcolumntype{x}[1]{>{\centering\arraybackslash}p{#1}}

\usepackage[utf8]{inputenc}
\usepackage[T1]{fontenc}

\usepackage[breaklinks]{hyperref}
\usepackage[capitalize]{cleveref}
\usepackage{CJKutf8}

\newcommand{\Tname}{Open-Set Text Recognition}
\newcommand{\tname}{open-set text recognition}
\newcommand{\textinset}{``in-set''}
\newcommand{\textoutofset}{``out-of-set''}
\newcommand{\textsideinfo}{side-information}

\newcommand{\textmykkcf}{Seen In-set Character}
\newcommand{\textmykucf}{Seen Out-of-set Character}

\newcommand{\textmyukcf}{Novel In-set Character}
\newcommand{\textmyuucf}{Novel  Out-of-set Character}

\newcommand{\textmyKKC}{SIC}
\newcommand{\textmyKUC}{SOC}
\newcommand{\textmyUKC}{NIC}
\newcommand{\textmyUUC}{NOC}

\newcommand{\mname}{label-to-prototype learning}
\newcommand{\Mname}{Label-to-Prototype Learning}
\newcommand{\textcharset}{character-set}
\definecolor{orange}{RGB}{220,100,0}
\definecolor{tenb}{RGB}{100,100,0}
\definecolor{ele}{RGB}{0,100,100}

\newcommand{\chgten}[1]{\textcolor{black}{#1}}
\newcommand{\chgtena}[1]{\textcolor{black}{#1}}

\newcommand{\chgele}[1]{\textcolor{black}{#1}}
\newcommand{\chgtwe}[1]{\textcolor{black}{#1}}
\newcommand{\cgxiii}[1]{\textcolor{black}{#1}}
\newcommand{\cgxiv}[1]{\textcolor{black}{#1}}
\newcommand{\cgxv}[1]{\textcolor{black}{#1}}
\newcommand{\cgxvi}[1]{\textcolor{black}{#1}}
\newcommand{\cgxvii}[1]{\textcolor{black}{#1}}
\newcommand{\cgxviii}[1]{\textcolor{black}{#1}}
\newcommand{\cgxix}[1]{\textcolor{black}{#1}}
\newcommand{\cgxx}[1]{\textcolor{black}{#1}}
\newcommand{\cgxxi}[1]{\textcolor{black}{#1}}
\newcommand{\cgxxii}[1]{\textcolor{black}{#1}}
\newcommand{\cgxxiii}[1]{\textcolor{black}{#1}}

\newcommand{\StSpace}[1]{\mathcal{#1}}
\newcommand{\StSet}[1]{\mathbf{#1}}
\newcommand{\Stfn}[1]{\mathrm{#1}}

\newcommand{\indexof}[1]{{[#1]}}
\newcommand{\variantof}[1]{{#1}}
\newcommand{\uprofit}[1]{#1_{m}}

\newcommand{\Nfeatdim}{d}

\newcommand{\Sfeatdim}{512}
\newcommand{\Tfeatdim}{const}
\newcommand{\Dfeatdim}{\# dimensions of the prototype space, 512 in this work}
\newcommand{\Afeatdim}{Ch.~\ref{sec:frame}}

\newcommand{\Nmaxtempbat}{b_{max}}

\newcommand{\Smaxtempbat}{512}
\newcommand{\Tmaxtempbat}{const}
\newcommand{\Dmaxtempbat}{The maximum  of the training process}
\newcommand{\Amaxtempbat}{Ch~\ref{sec:optim}}

\newcommand{\Nfracseen}{f_s}

\newcommand{\Sfracseen}{0.8}
\newcommand{\Tfracseen}{const}
\newcommand{\Dfracseen}{The fraction of seen characters for the label sampler}
\newcommand{\Afracseen}{Ch~\ref{sec:optim}}

\newcommand{\Nhardness}{h}

\newcommand{\Shardness}{0.5}
\newcommand{\Thardness}{const}
\newcommand{\Dhardness}{Limits extent of rectification the TPT module can do}
\newcommand{\Ahardness}{Ch~\ref{sec:tpt}}

\newcommand{\Nwemb}{\lambda_{emb}}

\newcommand{\Swemb}{0.3}
\newcommand{\Twemb}{const}
\newcommand{\Dwemb}{Weight of $\Nembloss$}
\newcommand{\Awemb}{Ch~\ref{sec:optim}}

\newcommand{\Nmaxitr}{n_{max}}

\newcommand{\Smaxitr}{-}
\newcommand{\Tmaxitr}{const}
\newcommand{\Dmaxitr}{The maximum iteration of the training process}
\newcommand{\Amaxitr}{Ch.\ref{sec:idet}}

\newcommand{\Nmaxtime}{l_\variantof{max}}

\newcommand{\Smaxtime}{-}
\newcommand{\Tmaxtime}{const}
\newcommand{\Dmaxtime}{The maximum sample length the model supports(including end of speech token).}
\newcommand{\Amaxtime}{Ch.~\ref{sec:idet}}

\newcommand{\Ntime}{l}

\newcommand{\Stime}{-}
\newcommand{\Ttime}{index}
\newcommand{\Dtime}{indicates a timestamp (the $\Ntime^{th}$ item) }
\newcommand{\Atime}{Ch~\ref{sec:tpt}}

\newcommand{\Ntimeupr}{\uprofit{l}}

\newcommand{\Stimeupr}{}
\newcommand{\Ttimeupr}{scalar}
\newcommand{\Dtimeupr}{the upper bound of timestamp length}
\newcommand{\Atimeupr}{Ch~\ref{sec:frame}}

\newcommand{\Nrec}{RE}
\newcommand{\Srec}{}
\newcommand{\Trec}{metric}
\newcommand{\Drec}{Recall of spotting samples containing $\Ncutest$.}
\newcommand{\Arec}{Ch.~\ref{sec:task_formulation}}

\newcommand{\Npre}{PR}
\newcommand{\Spre}{}
\newcommand{\Tpre}{metric}
\newcommand{\Dpre}{Precision of spotting samples containing $\Ncutest$.}
\newcommand{\Apre}{Ch.~\ref{sec:task_formulation}}

\newcommand{\Nfm}{FM}
\newcommand{\Sfm}{}
\newcommand{\Tfm}{metric}
\newcommand{\Dfm}{Fmeasure of spotting samples containing $\Ncutest$.}
\newcommand{\Afm}{Ch.~\ref{sec:task_formulation}}

\newcommand{\NLineACR}{LA}
\newcommand{\SLineACR}{}
\newcommand{\TLineACR}{metric}
\newcommand{\DLineACR}{Line accuracy}
\newcommand{\ALineACR}{Ch.~\ref{sec:task_formulation}}

\newcommand{\NCharACR}{CA}
\newcommand{\SCharACR}{}
\newcommand{\TCharACR}{metric}
\newcommand{\DCharACR}{Char accuracy}
\newcommand{\ACharACR}{Ch.~\ref{sec:task_formulation}}

\newcommand{\Nembloss}{L_{emb}}

\newcommand{\Sembloss}{}
\newcommand{\Tembloss}{loss}
\newcommand{\Dembloss}{The embedding loss pusing prototypes too close too close to each other away.}
\newcommand{\Aembloss}{Ch.\ref{sec:optim}}

\newcommand{\Nmodloss}{L_{model}}

\newcommand{\Smodloss}{-}
\newcommand{\Tmodloss}{loss}
\newcommand{\Dmodloss}{The total loss of the proposed method}
\newcommand{\Amodloss}{Ch.\ref{sec:optim}}

\newcommand{\Nclsloss}{L_{ce}}

\newcommand{\Sclsloss}{-}
\newcommand{\Tclsloss}{loss}
\newcommand{\Dclsloss}{The cross entropy of the classification task.}
\newcommand{\Aclsloss}{Ch.\ref{sec:optim}}

\newcommand{\Nradius}{s_{-}}

\newcommand{\Sradius}{-}
\newcommand{\Tradius}{scalar}
\newcommand{\Dradius}{Traiable decision boundary radius for all classes}
\newcommand{\Aradius}{Ch.\ref{sec:osp}}

\newcommand{\Ncharspace}{\StSpace{C}}

\newcommand{\Scharspace}{}
\newcommand{\Tcharspace}{space}
\newcommand{\Dcharspace}{Set of all characters.}
\newcommand{\Acharspace}{Ch.~\ref{sec:frame}}

\newcommand{\Ntempspace}{\StSpace{T}}

\newcommand{\Stempspace}{ \mathbb{R}^{32\times 32}}
\newcommand{\Ttempspace}{space}
\newcommand{\Dtempspace}{Template space containing $32 \times 32$ gray scale patches from the Noto-font (or other fonts). }
\newcommand{\Atempspace}{Ch.~\ref{sec:frame}}

\newcommand{\Nprotospace}{\StSpace{P}}

\newcommand{\Sprotospace}{\mathbb{R}^{\Nfeatdim{}}}
\newcommand{\Tprotospace}{space}
\newcommand{\Dprotospace}{Set of all characters.}
\newcommand{\Aprotospace}{Ch.~\ref{sec:frame}}

\newcommand{\Nctrain}{\StSet{C}_{train}}
\newcommand{\Sctrain}{}
\newcommand{\Tctrain}{set}
\newcommand{\Dctrain}{Set of distinct character(label) in the training set}
\newcommand{\Actrain}{Ch.~\ref{sec:task_formulation}}

\newcommand{\Nctest}{\StSet{C}_{test}}
\newcommand{\Sctest}{}
\newcommand{\Tctest}{set}
\newcommand{\Dctest}{Set of distinct character(label) in the testing set}
\newcommand{\Actest}{Ch.~\ref{sec:task_formulation}}

\newcommand{\Ncktest}{\StSet{C}^{i}_{test}}
\newcommand{\Tcktest}{set}
\newcommand{\Scktest}{}
\newcommand{\Dcktest}{ Labels in the testing set with side-information.}
\newcommand{\Acktest}{Ch.~\ref{sec:task_formulation}}

\newcommand{\Ncutest}{\StSet{C}^{o}_{test}}
\newcommand{\Scutest}{}
\newcommand{\Tcutest}{set}
\newcommand{\Dcutest}{Out-of-set labels in the testing set without side-information.}
\newcommand{\Acutest}{Ch.~\ref{sec:task_formulation}}

\newcommand{\Ncharlabel}{\StSet{C}_{label}}

\newcommand{\Scharlabel}{character}
\newcommand{\Tcharlabel}{set}
\newcommand{\Dcharlabel}{The set of characters appear in a batch of training data }
\newcommand{\Acharlabel}{Ch.\ref{sec:optim}}

\newcommand{\Ncharpos}{\StSet{C}_{pos}}

\newcommand{\Scharpos}{character}
\newcommand{\Tcharpos}{set}
\newcommand{\Dcharpos}{Sampled labels that appear in the batch of training data}
\newcommand{\Acharpos}{Ch.\ref{sec:optim}}

\newcommand{\Ncharneg}{\StSet{C}_{neg}}

\newcommand{\Scharneg}{character}
\newcommand{\Tcharneg}{set}
\newcommand{\Dcharneg}{Sampled labels that does not appear in the batch of training data}
\newcommand{\Acharneg}{Ch.\ref{sec:optim}}

\newcommand{\Ngt}{\StSet{Gt}}

\newcommand{\Sgt}{string}
\newcommand{\Tgt}{set}
\newcommand{\Dgt}{Label strings(ground truths) of samples a dataset.}
\newcommand{\Agt}{Ch.\ref{sec:task_formulation}}

\newcommand{\Npr}{\StSet{Pr}}

\newcommand{\Spr}{string}
\newcommand{\Tpr}{set}
\newcommand{\Dpr}{Prediction results of samples a dataset.}
\newcommand{\Apr}{Ch.\ref{sec:task_formulation}}

\newcommand{\Ncharbatch}{\StSet{C}_{batch}}

\newcommand{\Scharbatch}{character}
\newcommand{\Tcharbatch}{set}
\newcommand{\Dcharbatch}{The set of active labels in each training iteration, $\Ncharneg \cup \Ncharpos \cup\{[s],[-]\}$}
\newcommand{\Acharbatch}{Ch.\ref{sec:optim}}

\newcommand{\Nchar}{\StSet{C}}

\newcommand{\Schar}{character}
\newcommand{\Tchar}{set}
\newcommand{\Dchar}{An arbitary Set of characters.}
\newcommand{\Achar}{Ch.~\ref{sec:frame}}

\newcommand{\Nanytemps}[1]{\StSet{T}_{\indexof{#1}}}
\newcommand{\Ranytemps}[1]{}
\newcommand{\Sanytemps}[1]{\mathbb{R}^{32\times 32}}
\newcommand{\Tanytemps}[1]{set}
\newcommand{\Danytemps}[1]{All templates of the $#1^{th}$ character, each corresponds to a case. }
\newcommand{\Aanytemps}[1]{}

\newcommand{\Nproto}{P}

\newcommand{\Sproto}{\mathbb{R}^{\Nfeatdim{} \times {|T|}}}
\newcommand{\Tproto}{tensor}
\newcommand{\Dproto}{Prototypes for all classes. Note one class can be mapped to more than one prototypes depending on the number of cases.}
\newcommand{\Aproto}{Ch.~\ref{sec:frame}}

\newcommand{\Nanychar}[1]{c_\indexof{#1}}
\newcommand{\Ranychar}[1]{}
\newcommand{\Sanychar}[1]{}
\newcommand{\Tanychar}[1]{character}
\newcommand{\Danychar}[1]{A character in $\Ncharspace$}
\newcommand{\Aanychar}[1]{Ch\ref{sec:l2p}}

\newcommand{\Nanytemp}[1]{T_\indexof{#1}}
\newcommand{\Ranytemp}[1]{}
\newcommand{\Sanytemp}[1]{\mathbb{R}^{32\times 32}}
\newcommand{\Tanytemp}[1]{tensor}
\newcommand{\Danytemp}[1]{A template on the template space $\Ntempspace$. }
\newcommand{\Aanytemp}[1]{Ch\ref{sec:l2p}}

\newcommand{\Nanyproto}[1]{P_\indexof{#1}}
\newcommand{\Ranyproto}[1]{}
\newcommand{\Sanyproto}[1]{\mathbb{R}^\Nfeatdim}
\newcommand{\Tanyproto}[1]{tensor}
\newcommand{\Danyproto}[1]{A prototype of a ``case'' from a character.}
\newcommand{\Aanyproto}[1]{Ch\ref{sec:l2p}}

\newcommand{\Nfeatseq}{F}

\newcommand{\Sfeatseq}{\mathbb{R}^{\Ntime \times \Nfeatdim}}
\newcommand{\Tfeatseq}{tensor}
\newcommand{\Dfeatseq}{The feature of each character in one sample. $T$ indicates the length of the sample.}
\newcommand{\Afeatseq}{Ch\ref{sec:frame}}

\newcommand{\Nfeatmap}[1]{M_\variantof{#1}}
\newcommand{\Rfeatmap}[1]{}
\newcommand{\Sfeatmap}[1]{\mathbb{R}^{w \times h \times \Nfeatdim^{'}}}
\newcommand{\Tfeatmap}[1]{tensor}
\newcommand{\Dfeatmap}[1]{An (intermediate) feature map of the input word clip.}
\newcommand{\Afeatmap}[1]{Ch~\ref{sec:tpt}}

\newcommand{\Ndensx}{Dx}

\newcommand{\Sdensx}{\mathbb{R}_+^{w \times h}}
\newcommand{\Tdensx}{tensor}
\newcommand{\Ddensx}{The foreground density matrix for the $X$-axis.}
\newcommand{\Adensx}{Ch~\ref{sec:tpt}}

\newcommand{\Ndensy}{Dy}

\newcommand{\Sdensy}{\mathbb{R}_+^{w \times h}}
\newcommand{\Tdensy}{tensor}
\newcommand{\Ddensy}{The foreground density matrix for the $Y$-axis.}
\newcommand{\Adensy}{Ch~\ref{sec:tpt}}

\newcommand{\Nnewcord}{I}

\newcommand{\Snewcord}{\mathbb{R}_+^{w \times h \times 2}}
\newcommand{\Tnewcord}{tensor}
\newcommand{\Dnewcord}{The coordination mapping for the rectification.}
\newcommand{\Anewcord}{Ch~\ref{sec:tpt}}

\newcommand{\Nsamefn}{\Stfn{Same}}
\newcommand{\Ssamefn}{(str,str) \rightarrow \{0,1\}}
\newcommand{\Tsamefn}{function}
\newcommand{\Dsamefn}{return 1 if the two inputs are the same, 0 otherwise}
\newcommand{\Asamefn}{Ch.~\ref{sec:task_formulation}}

\newcommand{\NlTproto}{\Stfn{H}}

\newcommand{\SlTproto}{Char \rightarrow \{\mathbb{R}^{\Nfeatdim}\}}
\newcommand{\TlTproto}{function}
\newcommand{\DlTproto}{Mapping template to a normalized prototype.}
\newcommand{\AlTproto}{Ch~\ref{sec:l2p}}

\newcommand{\NlTtemp}{\Stfn{R}}

\newcommand{\SlTtemp}{Char \rightarrow {\mathbb{R}^{32\times 32}}}
\newcommand{\TlTtemp}{function}
\newcommand{\DlTtemp}{ Mapping a character label to all corresponding templates, each templates is a $32 \times 32$ gray scale patch of one case of the character.}
\newcommand{\AlTtemp}{Ch~\ref{sec:l2p}}

\newcommand{\NtTproto}{\Stfn{E}}

\newcommand{\StTproto}{\mathbb{R}^{32\times 32}\rightarrow \mathbb{R}^{\Nfeatdim}}
\newcommand{\TtTproto}{function}
\newcommand{\DtTproto}{Mapping template to a normalized prototype.}
\newcommand{\AtTproto}{Ch~\ref{sec:l2p}}

\newcommand{\Nfndensx}{\Stfn{Dx}}

\newcommand{\Nfndensy}{\Stfn{Dy}}

\newcommand{\Nfned}{\Stfn{ED}}

\newcommand{\Sfned}{(str,str) \rightarrow \mathbb{N}}
\newcommand{\Tfned}{function}
\newcommand{\Dfned}{Edit distance between strings.}
\newcommand{\Afned}{Ch~\ref{sec:task_formulation}}

\newcommand{\Nfnlen}{\Stfn{Len}}

\newcommand{\Sfnlen}{str \rightarrow \mathbb{N}^{+}}
\newcommand{\Tfnlen}{function}
\newcommand{\Dfnlen}{Returns the length of the string.}
\newcommand{\Afnlen}{Ch~\ref{sec:task_formulation}}

\newcommand{\Nfnrej}{\Stfn{Rej}}

\newcommand{\Sfnrej}{str \rightarrow \{0,1\}}
\newcommand{\Tfnrej}{function}
\newcommand{\Dfnrej}{Indicator function returns whether the string contains ``out-of-set'' characters in $\Ncutest$.}
\newcommand{\Afnrej}{Ch~\ref{sec:task_formulation}}

\newcommand{\Ncasesim}{\tilde{S}}

\newcommand{\Scasesim}{\mathbb{R}^{\Ntimeupr{} \times |\Nproto|}}
\newcommand{\Tcasesim}{tensor}
\newcommand{\Dcasesim}{Similarity of character feature at each timestamp $\Ntime$ and each \textbf{prototype}.}
\newcommand{\Acasesim}{Ch~\ref{sec:osp}}

\newcommand{\Nchasim}{\tilde{A}}

\newcommand{\Schasim}{ \mathbb{R}^{ \Ntimeupr{}\times (|\Nchar|+1)}}
\newcommand{\Tchasim}{tensor}
\newcommand{\Dchasim}{Similarity of character feature at each timestamp $\Ntime$ and each \textbf{character}.}
\newcommand{\Achasim}{Ch~\ref{sec:osp}}

\newcommand{\Nfinscr}{A}

\newcommand{\Sfinscr}{\mathbb{R}^{ \Ntimeupr{}\times (|\Nchar|+2)}}
\newcommand{\Tfinscr}{tensor}
\newcommand{\Dfinscr}{Final score of character feature at each timestamp (added the score for unknown character).}
\newcommand{\Afinscr}{Ch~\ref{sec:osp}}

\begin{document}

	\title{Towards \Tname{} via \Mname{}}

	\author[1]{Chang~Liu \fnref{fn1}}
	\ead{lasercat@gmx.us}
	\author[1]{	Chun~Yang \fnref{fn1}}
	\ead{chunyang@ustb.edu.cn}
	\author[1]{Hai-Bo~Qin}
	\ead{18701330507@163.com}
	\author[1]{Xiaobin Zhu}
	\ead{zhuxiaobin@ustb.edu.cn}
	\author[2]{Cheng-Lin Liu}
	\ead{liucl@nlpr.ia.ac.cn}
	\author[1]{Xu-Cheng~Yin\corref{cor1}}
	\ead{xuchengyin@ustb.edu.cn}
	\cortext[cor1]{Corresponding author}
	\fntext[fn1]{These authors contributed equally.}
	\address[1]{Department of Computer Science and Technology, School of Computer and Communication Engineering, University of Science and Technology Beijing, Beijing 100083, China }
	\address[2]{National Laboratory of Pattern Recognition, Institute of Automation, Chinese Academy of Sciences, Beijing 100190, China.}
	%
	%

 \begin{abstract}
	Scene text recognition is a popular topic and extensively used in the industry. Although many methods have achieved satisfactory performance for the close-set text recognition challenges, these methods lose feasibility in open-set scenarios, where collecting data or retraining models for novel characters \cgxxii{could yield a high cost}. \cgxx{For example, annotating samples for foreign languages can be expensive, whereas retraining the model each time when a ``novel'' character is discovered from historical documents costs both time and resources.} 
	In this paper, we introduce and formulate \cgxxii{a new} \tname{} task which demands the capability to \cgxx{spot and recognize} novel characters without retraining. \cgxxii{A label-to-prototype learning framework is also proposed as a baseline for the proposed task.} 
	\cgxxii{Specifically, the framework introduces a generalizable label-to-prototype mapping function to build prototypes (class centers) for both seen and unseen classes. An open-set predictor is then utilized to  recognize or reject samples according to the prototypes. The implementation of} rejection capability \cgxx{over out-of-set characters} allows \cgxxii{automatic spotting of unknown characters in the incoming data stream}.
	Extensive experiments show that our method achieves promising performance on a variety of zero-shot, close-set, and open-set text recognition datasets.
\end{abstract} 
\maketitle

\section{Introduction}
Text recognition is gaining popularity among both researchers and industry fellows due to its vast applications. Currently, many scene text recognition methods~\cite{Rosetta,ASTER,DAN} have achieved promising performance on the close-set text recognition benchmarks. However, most existing methods always fail to handle novel (unseen) characters that do not appear in the training set. 
\cgxvii{Specifically, these methods model the prototypes (centers) of classes as
	\cgxx{latent weights of a linear classifier. However, adding weights for novel characters will be difficult} without retraining the model. 
	This caveat makes conventional methods unfeasible \cgxx{as} collecting and annotating data can be very time-consuming and expensive. \cgxx{For example, the annotation of minority languages and ancient documents could take \cgxxii{a long time period and yield} high personnel costs.}	%
	Also, novel characters may be continuously discovered from the incoming data stream, especially \cgxxiii{when processing historical documents or internet oriented images. In such cases, retraining the model each time a novel character is found would yield considerable time and resource costs.} 
} 
In the literature, a few text recognition methods are capable to handle novel characters. These methods can be divided into two different schemes. \cgxxii{One category~\cite{taktak,fewran,zhang20pr} exploits the radical composition of each character, consequentially limiting most of them to} the Chinese language.
Another category is based on weight imprinting~\cite{imp}, which generates prototypes with corresponding glyphs~\cite{cm19,eccv20,eccvfork}. 
Despite showing no language-specific limitation, many \cgxx{do} not scale well on large label sets due to the \cgxxiii{training} cost of the \cgxiii{imprinting module}.
\cgxxii{For both categories, despite a few methods possessing text-line recognition capability~\cite{zhang20pr,jinic21,eccv20}, very few  demonstrate competitive performance on standard close-set benchmarks~\cite{www}, limiting their feasibility in practice.}

\cgxx{Furthermore}, the aforementioned zero-shot text recognition methods are not capable to reject \textoutofset{} samples like Open Set Recognition (OSR) methods~\cite{osr-survey,osb16}.
\label{rev:hitl}
\cgxxii{The rejection capability can help the users to quickly spot newly emerging characters in the incoming data stream, whereas the recognition capability enables a quick adaption without retraining the model. Combining these two capabilities yields a human-in-the-loop recognition system capable to evolve with the input data, which we formulate as the \tname{} task.}
\cgxx{Here, we propose a label-to-prototype learning framework to efficiently address \cgxxii{this} new task in a language-agnostic manner, while preserving competitive performance on close-set benchmarks.} 
\cgxx{To dynamically build prototypes for both seen and unseen (novel) characters, a label-to-prototype module is proposed to map each character to its corresponding prototypes (classifier weights) by encoding glyphs of all its cases. The module augments the conventional text recognition framework~\cite{www} as an extra stage.}
\cgxxii{The open-set predictor is then introduced to categorize or reject character features at each-time stamp according to its similarities with the dynamically generated prototypes, replacing the linear classifier in the Pred. stage~\cite{www}.}
Moreover, we propose a lightweight topology preserving rectification module to handle skewed samples. The module applies \cgxxiii{a mesh-grid transformation to intermediate feature maps, with parameter estimation fused with the backbone} avoiding \cgxxii{dedicated rectification networks like~\cite{www,moran,scrn}}. 
A variety of quantitative and qualitative experiments show that our proposed framework achieves promising performance on all open-set, zero-shot, and close-set text recognition datasets.

In summary, our main contributions are: 
\begin{itemize}
	\item Formulating the \tname{} task that demands further capabilities to spot novel characters and adapt to recognize them without retraining the model.   
	\item Proposing a \mname{} framework capable to address novel characters while  keeping reasonable speed and competitive performance on conventional benchmarks.%
	\item Proposing a topology-preserving transformation technique for text recognition \cgxxii{by performing} fast fine-grained rectification alongside the feature extraction.	
	
\end{itemize}  

\section{\Tname{}}
\subsection{Task Definition}
\label{rev:task_formulation}
\label{sec:task_formulation}

\begin{figure}[t]
	\centering
	\includegraphics[width=0.75\linewidth]{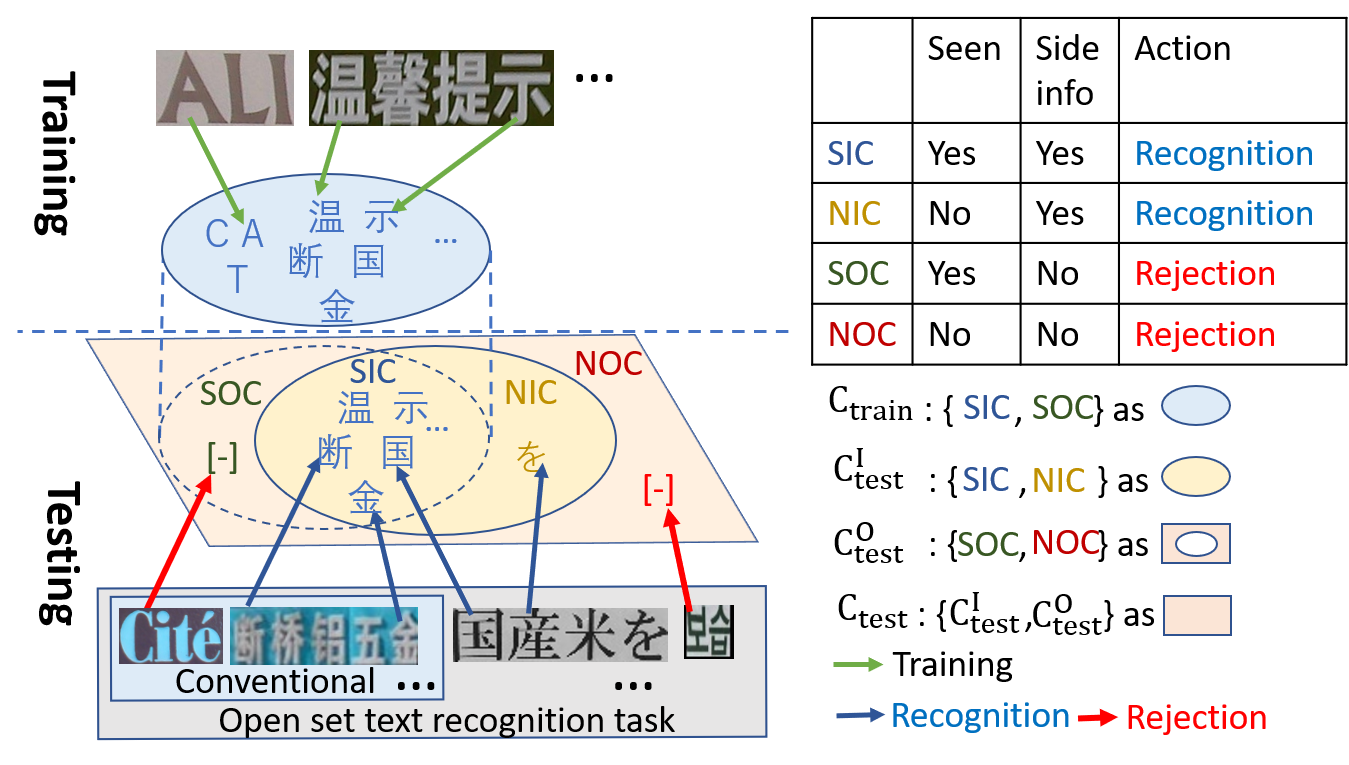}
	\caption{Open-set text recognition task: During the evaluation process, some samples may include characters not appearing in the training set. \cgxxii{The task requires the model to recognize both seen and novel characters in $\Ncktest$ (blue arrow) and rejecting samples with ``out-of-set'' characters in $\Ncutest$ (red arrow). }
	}
	\label{fig:oops}
\end{figure}
\cgxxii{Generally speaking, the \tname{} task (OSTR, illustrated in Fig.~\ref{fig:oops}) extracts text from images that potentially contain novel characters unseen in the training set. \cgxxiii{Specifically, a} character $\Nanychar{i}$ is considered ``novel'' if $\Nanychar{i} \notin \Nctrain{}$, where  $\Nctrain{}$ is the set of all characters that appear in the training samples. 
	\cgxxiii{Likewise, characters of the testing set form the testing character set $\Nctest{}$.  $\Nctest{}$} can be disjointly divided into $\Ncktest$ for \textinset{} characters \cgxxiii{with corresponding \textsideinfo{} provided during evaluation}, and $\Ncutest{}$ for \textoutofset{} characters without \textsideinfo{}. Here, \textsideinfo{} refers to label information that uniquely defines a character, e.g., glyphs~\cite{eccv20}, component composition~\cite{hde,taktak}, or other equivalent information.
	The task requires recognizing characters in $\Ncktest$ w.r.t. the provided \textsideinfo{} and reject the characters from $\Ncutest{}$ by predicting them as the ``unknown'' label `[-]'.} 

\cgxxiii{Note in applications, $\Nctest$ refers to all characters in the entire data stream. Hence in many cases  users have no prior knowledge of novel characters at the beginning. %
However, when novel characters are spotted by inspecting rejected samples, the user can move them to $\Ncktest$ for future recognition, by providing their \textsideinfo{} to the model. Vice versa, the user may move an obsolete character from  $\Ncktest$ to  $\Ncutest{}$ by removing its \textsideinfo{}.  
Hence, $\Ncktest$ can be volatile} as a result of users adding or removing characters.
\cgxxii{
	For addressing convenience, we accordingly divide characters to 4 categories as shown in Figure~\ref{fig:oops}: ``\textmykkcf~(\textmyKKC)'', ``\textmykucf~(\textmyKUC)'', ``\textmyukcf~(\textmyUKC)'', and the ``\textmyuucf~(\textmyUUC)''. \cgxxiii{The first term indicates whether the character is ``seen'' or ``novel'' to training samples, and the second tells whether it is in $\Ncktest$ with its \textsideinfo{} provided for evaluation.}
}

\cgxxii{Metric-wise, two popular metrics in the text recognition community are used to evaluate the recognition performance.} One is the line accuracy(\NLineACR{})~\footnote{The line accuracy is also known as word accuracy (WA), same to 1-WER (Word Error Rate).}~\cite{www}, defined as,
\begin{equation}
	\NLineACR{}=\frac{1}{N}\sum_i^N \Nsamefn(\Ngt_\indexof{i},\Npr_\indexof{i}),
\end{equation}
where $N$ is the size of the dataset, $\Ngt_\indexof{i}$  and $\Npr_\indexof{i}$ correspondingly indicate the label and the prediction of the $i^{th}$ sample. 
Another is \cgxxi{the} character accuracy  (\NCharACR)~\footnote{The character accuracy is also known as accuracy rate (AR), same to 1-NED (normalized edit distance).}~\cite{art},
\begin{equation}
	\NCharACR=1-\frac{\sum_i^N \Nfned(\Npr_\indexof{i},\Ngt_\indexof{i})}{\sum_i^N \Nfnlen(\Ngt_\indexof{i})}.
\end{equation}

\cgxxii{The capability to reject characters in $\Ncutest{}$ is measured with word-level recall~($\Nrec$), precision ($\Npre$), and F-measure ($\Nfm$).}
\begin{equation}
	\Nrec=\frac{\sum_i^N \Nfnrej(\Npr_\indexof{i})}{\sum_i^N \Nfnrej(\Ngt_\indexof{i})}, \
	\Npre=\frac{\sum_i^N \Nfnrej(\Npr_\indexof{i})\Nfnrej(\Ngt_\indexof{i})}{\sum_i^N \Nfnrej(\Npr_\indexof{i})}.\
\end{equation}
\cgxxii{Here, $\Nfnrej$ is an indicator function that returns whether the string contains ``out-of-set'' characters (\textmyKUC{} and \textmyUUC{}).}
	\cgxxiii{Since word samples containing out-of-set characters~(predicted as ``unknown'') require human inspection whether ``in-set'' characters appear or not, we use word-level measurements for simplicity.}
	\cgxxii{Specifically, the recall describes the likelihood for the system to reject a sample containing \textoutofset{} characters and alert the user to adjust $\Ncktest$. The precision measures the ratio of fake positive warnings needing manual dismissal. 
	Finally, the F-measure is given as an overall metric of recall and precision,}
\begin{equation}
	\Nfm=\frac{2\Nrec*\Npre}{\Nrec+\Npre}.
\end{equation}

\subsection{Dataset}
\label{sec:ossds}
In this work, we construct an open-set text recognition dataset with a collection of existing \cgxx{openly available} datasets.
Here, we choose Chinese scene text images for training while Japanese scene text images for evaluation, due to the abundant amount of samples that can be found in these two languages.
\cgxx{The list of upstream datasets and the processing scripts} are also released alongside our codes\footnote{\url{https://github.com/lancercat/OSOCR}}. 
\cgxvii{ Specifically, the training set contains Chinese and Latin samples collected from  ART \cite{art}, RCTW \cite{rctw}, LSVT \cite{lsvt}, CTW \cite{ctw}, and the Latin-Chinese subset of the MLT dataset~\footnote{Crops with language annotated as Latin and Chinese.}.  The \textcharset{} $\Nctrain{}$ contains  \cgxx{Tier-1 simplified} Chinese characters, English letters(`A'-`z'), and digits (`0'-`9'). Samples with characters not covered by $\Nctrain{}$ are excluded from the training set.}
The testing dataset contains Japanese samples drawn from the MLT dataset. 
\cgxx{The \textcharset{} $\Nctest{}$ contains all $1460$ characters that appear in this subset.} 

The evaluation is performed under \cgxxii{five setups using different splits over $\Nctest$. The first setup is closer to the generalized zero-shot learning tasks (GZSL) where all $1460$ characters go to $\Ncktest{}$. The second splits all novel characters to $\Ncutest{}$ to match the open-set recognition (OSR) setup. The third introduces \textmyKUC{} to the second setup. The fourth split the Hiragana and Katakana into $\Ncutest{}$, imitating the GOSR setup. Finally, the fifth introduces \textmyKUC{} to $\Ncutest{}$ of task four, implementing the full OSTR task.}
To simplify the task and focus on the open-set problem, we remove all vertical texts in the training and testing sets. \cgxxiii{The testing set contains a total of} $4009$ text lines in the testing set.
Note that because the \textcharset{}s of Japanese and Chinese overlap, the conventional close-set methods do not yield a zero accuracy.

 \section{Related Works}
\subsection{Relations with Other Tasks}
The open-set text recognition (OSTR) task can be regarded as a combination of the open-set recognition task(OSR)~\cite{tosr,liucls,liuosr} and the GZSL task~\cite{gzsl-survey}.
\cgxxii{
Specifically, the OSTR requires recognizing both seen and novel characters in $\Ncktest$ with side-information like GZSL tasks~\cite{gzsl-survey}. Furthermore, the OSTR requires the rejection capability on samples from classes not covered by $\Ncktest$,  seen or unseen. Despite the subjects for rejection being slightly different, Both OSTR and OSR imply closed decision boundaries for each class to recognize. 
In summary, compared to the GZSL tasks, the OSTR task further adds the requirement on rejection capability, which enables actively spotting samples from ``out-of-set'' classes (not in $\Ncktest$) in applications.  Compared with the OSR task, besides the capability to spot samples from novel classes from the data stream, the OSTR introduced the capability to recognize samples from novel classes \cgxxiii{once} the corresponding side-information provided, allowing fast and incremental model adaption without retraining the model. 
 Noteworthy, the recognition capability of novel classes with closed decision boundaries implies that the samples of each novel class shall gather closely to its class center(s). Since this ``gather'' property is not affected by whether the side-information is known to the model, ``cognition'' on ``unknown unknown classes''~\cite{osr-survey} is implemented in the OSTR task. Hence, the OSTR task can be considered as a variant of the generalized open-set  text recognition as well. 
}
\begin{table}	
	\caption{
		Comparison with typical text recognition tasks. 
	}
	\begin{center}
		\resizebox{\linewidth}{!}{
			\begin{tabular}{c|p{4cm}|p{4cm}|p{4cm}}
				\hline
				Task& Close-set Text Recognition \cite{www,Rosetta,DAN,ASTER} & Zero-shot Chinese Character Recognition \cite{cm19,hde,fewran}  & Open-set Text Recognition (OSTR)\\
				\hline
				Input& Word &Character& Word \\
				\hline
				\textmyKKC &Recognition&No&Recognition\\
				\hline
				\textmyKUC &No&No&Rejection\\
				\hline
				\textmyUKC &No&Recognition&Recognition\\
				\hline
				\textmyUUC &No&No&Rejection\\
				\hline
				Language &Unlimited&Chinese&Unlimited\\
				\hline
		\end{tabular}}
		\label{tab:taskcmp}
	\end{center}
\end{table}

\cgxxii{
The proposed task is also related to several existing text recognition tasks as summarized in Table~\ref{tab:taskcmp}.
Compared to the close-set text recognition task solely focusing on the \textmyKKC{}, the OSTR task relaxes the $\Nctest{} \subseteq \Nctrain{}$  assumption by introducing novel labels which do not appear in the training samples.  On the other hand, compared to the zero-shot character recognition task~\cite{hde,fewran} which can also be regarded as a special case of OSTR when $\Nctest{} \cap \Nctrain{} = \varnothing$, $\Ncutest{}=\varnothing$, and all samples have a constant length of $1$. 
Furthermore, compared to both tasks, the OSTR task introduces the rejection of out-of-set characters, allowing two folds of capability. First, the rejection capability would allow the user to get a notification when novel characters appear in the data stream. Second, it allows the user to optionally skip recognition of rare characters to speed up, while keeping track of the rejected samples, which may include the skipped characters, for later checking.
} 
\subsection{Close-Set Text Recognition}
Most conventional text recognition methods can fit into the four-stage \cgxxi{framework}~\cite{www}, i.e., transformation (Trans.), feature extraction (Feat.), sequence modeling (Seq.), and prediction (Pred.). The Trans. stage \cgxxiii{rectifies the input image, which can be omitted} in some methods~\cite{DAN,CRNN}. In~\cite{www} and~\cite{RARE}, a spatial transformer network is used for automatic rectification, other methods~\cite{scrn,moran} construct heavier neural networks and \cgxx{implement} more complex transforms. 
The Feat. stage generally uses a convolutional network \cgxx{to extract visual features from the rectified images, e.g., ResNet~\cite{Resnet}, RCNN~\cite{RCNN}, and VGG~\cite{VGG}.} 
\cgxix{
The Seq. stage captures the contextual information of the whole sequence, by using LSTM layers~\cite{AON,CRNN,DAN}, transformers~\cite{satrn}, or simply nothing~\cite{Rosetta}.}
\cgxx{The Pred. Stage decodes the feature map into predictions.} Typical implementations can be divided  into two categories according to whether the  segmentation is explicitly supervised. \cgxxiii{Specifically, implicitly supervised segmentation} include RNN-Attention decoders~\cite{SAR}, CTC~\cite{CRNN}, and decoupled attention decoders~\cite{DAN}.  \cgxxi{On the other hand, explicit segmentation methods generally implement this stage with semantic segmentation heads, e.g.,  TextSpotter~\cite{ts} and CA-FCN~\cite{cafcn}.}
In this stage, a linear classifier is adopted by most \cgxx{implementations} to decode the \cgxxi{segmented} feature at each timestamp \cgxxi{into character prediction}. The linear classifier, in a more general way to put it, can be treated as a ``character classifier'' as it decodes its input into ``characters''. \footnote{Depending on the methods, the ``characters'' may include special tokens, e.g., $\epsilon$ in CTC based methods and ``EOS'' for RNN-Attention based methods.}  

\cgxxi{Due to the latent nature of weights in the linear predictor, constructing corresponding weights for a novel label can be difficult without retraining. Here, our proposed framework introduces a label-to-prototype module as an extra stage to construct prototypes for both novel and seen characters.}
\cgxx{An} open-set prediction module is \cgxx{also} proposed to replace the linear classifier used in the Pred. stage offering rejection capability of \textoutofset{} characters including \cgxxii{\textmyUUC{} and \textmyKUC{} characters}.

\subsection{Zero-Shot Text Recognition}
Most current zero-shot text recognition methods focus on Chinese Character recognition due to its \cgxx{challenging} large character-set. However, many methods~\cite{fewran,hde,jinic21,taktak} require detailed annotations on the \cgxx{radical composition} of each character (mainly the radical composition tree or stroke sequence of a character). Such knowledge is mostly used at the decoder-side and implemented as an RNN which is hard to parallelize~\cite{jinic21}. Some recent methods are seen to use it on \cgxix{the encoder side}~\cite{cm19,hde,jinic21}, which can avoid the RNN radical decoding process during evaluation.  
A few radical-based methods like  \cite{jinic21,zhang20pr} are capable to  perform Chinese text line recognition tasks on private datasets. 
However, mapping labels to corresponding component sequences requires strong domain knowledge and the annotating process is also a tedious job. \chgtwe{Furthermore, text components are mainly specific to Chinese characters, which practically limits the feasibility in multi-language scenarios.}
Recently, Zhang et al.~\cite{eccv20} proposed a visual-matching-based method that can handle novel characters in text lines without language limitations.
However, this method would yield \cgxxi{a significant} computation burden during training, \cgxxi{for} encoding a large ``glyph-line image'' caused by the large charset. Also, this method shows very limited performance on conventional close-set benchmarks, which renders it less feasible for real-world applications.  

Our proposed method only takes templates from the Noto fonts and does not need any domain knowledge of the specific language.
Furthermore, the rejection capability, vital to active novel characters spotting, is missing for most methods. Hence, these methods cannot be considered full open-set recognition methods.
	Currently, zero-shot text recognition methods are mostly benchmarked on property datasets~\cite{jinic21,zhang20pr}, or datasets with a limited number of \cgxiii{distinctive} classes~\cite{eccv20}. Thus, we provide an openly accessible open-set benchmark dataset in this work.%

\begin{figure*}[!t]
	\centering
	\includegraphics[width=0.9\linewidth]{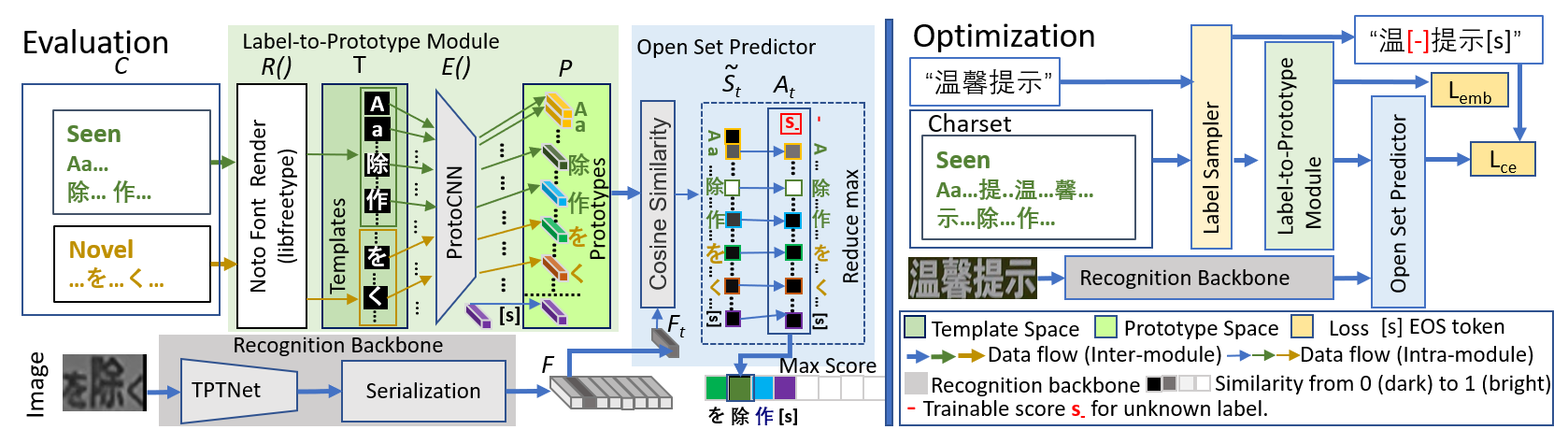}
	\caption{
		\cgxiv{
		Evaluation phase and \cgxvi{optimization} phase of our proposed framework. During evaluation, novel and seen characters are mapped corresponding prototypes via the label-to-prototype module, which is cached for all evaluation samples. \cgxiv{ During optimization, a label sampler is introduced to produce unknown labels by mini-batching the seen characters. The unknown labels ``[-]'' are used to train the rejection threshold $s_-$  for ``out-of-set'' characters.}
		Image feature is then extracted and serialized with the text recognition backbone.} Finally, the open-set predictor computes the similarity $\Ncasesim_\Ntime$ between the extracted feature $\Nfeatseq_{\Ntime}$ and prototypes $\Nproto$, and then reduces to the character score vector $\Nfinscr_{\Ntime}$.
		}
	\label{fig:ioocr}
\end{figure*} \section{Proposed Framework}

\subsection{Framework Overview}
\label{rev:framework_novelty}
\label{sec:frame}
\cgxxii{
	In this work, we propose a \mname{} framework (Fig.~\ref{fig:ioocr}) as a baseline of the proposed OSTR task, while retaining competitive performance on standard close-set~\cite{www} and zero-shot Chinese character~\cite{hde} benchmarks. 
	The proposed framework is composed of four main parts, namely the Recognition Backbone (Sec.~\ref{sec:tpt}), the 
	Label-to-prototype Module (Sec.~\ref{sec:l2p}), the Open-set-Predictor (Sec.~\ref{sec:osp}), and the Label Sampler (Sec.~\ref{sec:optim}).
}	
\cgxxii{
	Specifically, the Recognition Backbone locates, sorts, and extracts visual features $\Nfeatseq{}: \Sfeatseq$ of each character in the input sample. Most conventional text recognition methods can be used for this part after removing the linear classifier.\footnote{Methods with a recurrent decoder~\cite{SAR,nrtr} need an extra change on the ``hidden state'' to remove the dependency on history predictions.}}
\cgxxii{
	The \mname{} module provides a tractable mapping  $\NlTproto{}$ from side information of a character label to its corresponding prototypes $\Nproto$, each defining the center of a corresponding ``case'' of the  character. E.g., $\NlTproto{}($`a'$)$ returns two prototypes, one for uppercase `A' and another for lowercase `a'.
	The Open-set Predictor is then used to categorize or reject each input character features $\Nfeatseq_\indexof{l}$ according to the decision boundaries, defined with corresponding prototypes and the globally shared radius $\Nradius$.}
\cgxxii{Finally, the Label Sampler is proposed for the training phase to produce out-of-set samples and reduce the training burden of the \mname{} module by sampling a reasonable portion of labels from all training characters  $\Nctrain$ at each iteration.}
\subsection{Topology-Preserving Transformation Network}
\label{sec:tpt}
\cgxxii{
For the Recognition Backbone, we implement a Topology-Preserving Transformation Network (TPTNet) adapted from the Decoupled Attention Network (DAN).~\cite{DAN} Like DAN, the TPTNet first extracts visual feature $M$ with \cgxxiii{a ResNet backbone}, then a Concurrent Attention Module (CAM) is adopted to sample the character feature sequence $\Nfeatseq:({\Nfeatseq_{1},...,\Nfeatseq_{\Ntime},...,\Nfeatseq_{\Ntimeupr}})$ from $M$.  The RNN used to model the contextual information is removed for speed-performance trade-off.
} 

\begin{figure}[t]
	\centering
	\includegraphics[width=0.6\linewidth]{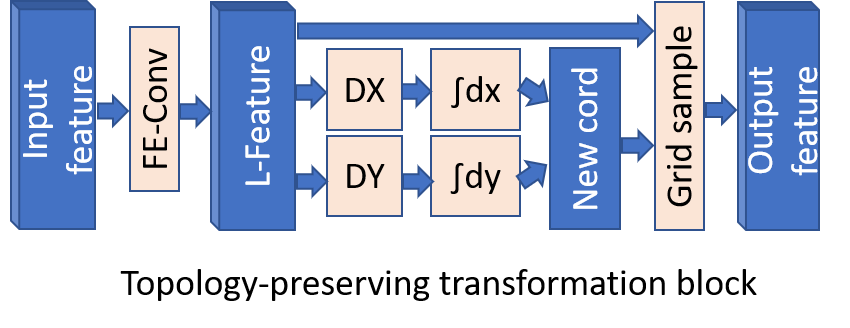}
	\caption{
		Topology-preserving transformation: Each input feature map is first processed into the local feature map $M_l$ with a convolutional block (FE-Conv). Next, we estimate the effective feature density with $DX$ and $DY$ modules. Then, we integrate the density into destination coordinates. Finally, we rectify $M_l$ with a grid sampler according to the destination coordinates.
	}
	\label{fig:lensb}
\end{figure}
\cgxxii{In this work, a fused-in feature-level mesh grid rectification is proposed to alleviate the common vulnerability of perspective transformations and various irregular distortions.}
The fused block~(Fig.~\ref{fig:lensb}) takes a feature map  $\Nfeatmap{i}$ as input, and produces a processed and rectified feature map $M_o$ as output. Specifically, the feature map is first processed into a local feature map $\Nfeatmap{l}$ with a convolutional block,
\begin{equation}
	M_l=\Stfn{Conv}(\Nfeatmap{i}).
\end{equation}
Next, we estimate the feature density $\Ndensx~\in~\Sdensx$ on the $X$ direction and $\Ndensy~\in~\Sdensy$ on the $Y$ direction with corresponding functions $\Nfndensx{}$ and $\Nfndensy{}$,
\begin{equation}
	\begin{split}
		Dx =& \Stfn{Sigmoid}(\Stfn{FC_x}(\Nfeatmap{l}))+b :=\Nfndensx(\Nfeatmap{l}), \\
		Dy =& \Stfn{Sigmoid}(\Stfn{FC_y}(\Nfeatmap{l}))+b :=\Nfndensy(\Nfeatmap{l}),\\
	\end{split}
\end{equation}
\cgxxii{where $\Stfn{FC}$ indicates $1\times1$ convolution layers and }$\Nhardness$ is a constant number controlling the extent of rectification \chgtwe{that} the module can perform. The densities $\Ndensx$ and $\Ndensy$ are \cgxx{then} integrated and normalized into the \cgxxii{coordination mapping $\Nnewcord \in \Snewcord$},
\begin{equation}
		\Nnewcord_\indexof{h,w} = [w\frac{\sum_{i=0}^{w}\Ndensx_\indexof{h,i}}{\sum_{i=0}^{\uprofit{w}}\Ndensx_\indexof{h,i}}, h\frac{\sum_{j=0}^{h}\Ndensy_\indexof{j,w}}{\sum_{j=0}^{\uprofit{h}}\Ndensy_\indexof{j,w}}].
\end{equation}
\cgxxii{Specifically, $\Nnewcord$  maps the coordination from the rectified feature map $\Nfeatmap{o}$ to the source feature map $\Nfeatmap{l}$, and the mesh-grid transformation is implemented by sampling $\Nfeatmap{l}$ into $\Nfeatmap{o}$ according to $\Nnewcord$ with a linear sampler.}
As the density is always greater than zero, the integration is able to preserve the input topology, i.e, lines without intersection will not intersect after the transformation and vice-versa.  The gradients of coordinates are computed \cgxvii{similarly} to the deformable convolution network in~\cite{deform}. %
\cgxxi{The estimation of transformation parameters $\Nnewcord$ shares most computation with the feature \cgxxii{extraction, avoiding} costly dedicated rectification networks in conventional approaches~\cite{moran,scrn}.} 
\subsection{\Mname{} Module}
\label{sec:l2p}
\cgxxii{The key point of open-set text recognition is to construct a tractable mapping $\NlTproto{}$ from the label space $\Ncharspace$ to the prototype space $\Nprotospace$, which can be applied to both seen and novel characters. Note $\NlTproto$ could  potentially  be a one-to-many mapping as a character may have multiple cases.}
\cgxxii{In this work, we propose the \mname{} module decomposing $\NlTproto{}$ into two mappings, namely $\NlTtemp$ and $\NtTproto{}$, by introducing} the intermediate template space $\Ntempspace:\Stempspace$  containing  glyphs drawn from the Noto-font.~\footnote{\chgten{The Noto fonts for most languages are available} at \url{https://noto-website-2.storage.googleapis.com/pkgs/Noto-unhinted.zip}.}

\cgxxi{
	Here, $\NlTtemp: \Ncharspace \rightarrow \Ntempspace$ maps each character in the character set $\Ncharspace$ to the glyphs of all its cases}. ``Case'' here refers to different forms of a character, e.g., upper-case and lower-case, simplified Chinese and traditional Chinese, Hiragana and Katakana, and different contextual shapes as in Arabic.  As \cgxxi{the} Noto-font covers most characters in many languages \cgxix{with a uniform style}, $\NlTtemp$ is highly likely to generalize to novel characters not appearing in the training set. \cgxix{Also,} $\NlTtemp$ requires little \cgxxii{domain knowledge of the specific languages, e.g. , part information and composition information for each character.}

\cgxxi{$\NtTproto: \Ntempspace \rightarrow \Nprotospace$ is the mapping from the template space to the prototype space. The mapping is implemented with the ProtoCNN module, \cgxxii{ which includes a ResNet18 network, a normalization layer, and a trainable latent prototype $\Nanyproto{s}$ for the end-of-speech ``character'' (`[s]')}. 
	Prototypes of normal characters are generated by encoding and normalizing the corresponding glyphs with the ResNet,}
\begin{equation}
	\begin{split}
		\Nanyproto{j}&=\frac{Res18(\Nanytemp{j})}{|Res18(\Nanytemp{j})|}.\\
	\end{split}	
\end{equation}
The normalization is conducted to alleviate the effect of the potential bias of character frequency between the training set and the testing set.
For $\NtTproto$ to be \cgxxii{generalizable} to novel classes, the templates of seen characters need to fill the template space $\Ntempspace$ densely enough, which requires the framework to be trained on languages with large \textcharset{}s. \cgxxii{As the normalized prototypes can be cached before evaluation and updated incrementally on \textcharset{} change, the evaluation-time overhead is negligible}.

\subsection{Open-Set Predictor}
\label{sec:osp}
In this work, an open-set predictor is proposed to replace the linear classifiers used \cgxxii{in} conventional text recognition methods~\cite{www}.
Like the \cgxxi{classifiers} in the weight imprinting \cgxx{methods}~\cite{imp}, \cgxxi{
	the \cgxxii{open-set predictor performs classification for characters in $\Ncktest$} by comparing the generated prototypes to visual features.}
The module further adopts closed decision boundaries~\cite{osb16,osb21} to achieve rejection on \textoutofset{} characters (in $\Ncutest$) \cgxx{not similar to any provided prototypes.}   %

\cgxxii{Specifically, the open-set predictor} first computes the \cgxvii{``case-specific''} similarity scores $\Ncasesim{} \in \Scasesim{}$ via  the scaled product of the visual feature $\Nfeatseq \in \Sfeatseq$ and the normalized prototype matrix $\Nproto\in \Sproto$,
\begin{equation}
	\Ncasesim{}=\alpha \Nfeatseq{}\Nproto.
\end{equation}
Here, the similarity score of the feature at timestamp $t$ and the $n^{th}$ prototype can be written as,
\begin{equation}
	\label{eq:WCWC}
	\begin{split}
		\Ncasesim{}_\indexof{\Ntime{},j} =&\alpha \Nfeatseq{}_\indexof{\Ntime{}} \Nproto_\indexof{j} =\alpha|\Nproto_\indexof{j}||\Nfeatseq{}_\indexof{\Ntime{}}| \Stfn{cos}(\Nproto_\indexof{j},\Nfeatseq{}_\indexof{\Ntime{}})\\
		=& \alpha|\Nfeatseq{}_\indexof{\Ntime{}}| \Stfn{cos}(\Nproto_\indexof{j},\Nfeatseq{}_\indexof{\Ntime{}}), ~~\forall j~|\Nproto_\indexof{j}|=1.\\
	\end{split}
\end{equation} 
Since $\alpha|\Nfeatseq{}_\indexof{\Ntime{}}|$ is a constant number given a certain timestamp $\Ntime{}$, 
$\Ncasesim{}_\indexof{\Ntime}$ can be interpreted as \cgxxi{the} scaled cosine similarities between all prototypes in $\Nproto$ and \cgxv{the} visual feature $\Nfeatseq_\indexof{\Ntime{}}$. 
Thus, $\Ncasesim{}$ is detonated as the similarity score. 
Due to the exponential operator in the Softmax function, $\alpha|\Nfeatseq{}_\indexof{\Ntime}|$ controls how \cgxxii{much} the predicted probability would be close to one-hot. \cgxx{Hence, $|\Nfeatseq_\indexof{\Ntime}|$ is interpreted as the confidence of timestamp $\Ntime$, and $\alpha$ interprets as the overall model confidence.  }
Also, since $|\Nanyproto{j}|$ can be interpreted as the overall ``preference'' of prototype $j$, the prototypes are normalized in the label-to-prototype module to \cgxx{alleviate} potential frequency-related bias. 

The module then applies max reduction on case-wise similarity $\Ncasesim{}$, \cgxxii{producing} the label-wise classification scores $\Nchasim \in \Schasim$, i.e.,
\begin{equation}
	\Nchasim_\indexof{\Ntime,i}= \max_{\{j|\Stfn{\phi}(j)=i\}}(\Ncasesim{}_\indexof{\Ntime,j}),
\end{equation}
where $\Stfn{\phi}$ is the label function that \cgxxiii{reduces} all $|\Nproto|$ prototypes to $|\Nchar|+1$ labels, and the extra \cgxxii{one} comes from the \cgxvii{``end-of-speech''} token. Specifically,  $\Stfn{\phi}(j)$ returns the character which has the $j^{th}$ template as one of its cases.
Combining this strategy with the mapping operation in the label-to-prototype module allows us to train the framework with case agnostic annotations, without explicitly aligning cases with drastic appearance differences to the same region on \cgxx{the} prototype space $\Nprotospace$, e.g., 'A' and 'a'.\label{rev:rejection}
\cgxxii{
The module rejects samples not similar to any provided prototypes by predicting them into the \textoutofset{} label `[-]'. 
The  \textoutofset{} label is not associated with a dedicated prototype as it covers many characters that do not share common visual traits.
Instead, the module uses a trainable score $\Nradius$  attached to $\Ncasesim_{\Ntime{}}$ as the similarity score for the \textoutofset{} label at each timestamp $\Ntime$, } 
yielding the final score vector $\Nfinscr \in \Sfinscr$,
\begin{equation}
	\Nfinscr_\indexof{\Ntime}=[\Nchasim_\indexof{\Ntime,1},\Nchasim_\indexof{\Ntime,2},...,\Nchasim_\indexof{\Ntime,|\Nchar|+1},\Nradius].
\end{equation}
\cgxxii{During training, the classification loss pushes $s_{-}$ above other scores on \textoutofset{}
	 characters, for ``in-set'' characters,  $s_{-}$ is pushed below the score of correct classes.} \cgxxiii{During evaluation, \textoutofset{} characters would yield similarities less than $s_{-}$ to any prototypes due to the shape difference, yielding a rejection. Hence, $s_{-}$ can also be interpreted as the radius of the decision boundary~\cite{osb16}, or a similarity threshold.}

\begin{figure}[t]
	\centering
	\includegraphics[width=0.6\linewidth]{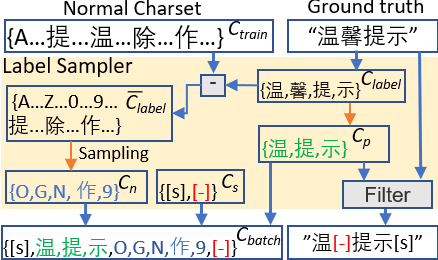}
	\caption{
		The \cgxxii{proposed label sampler}. The module samples a subset of the seen characters according to the labels for each training iteration.
	}
	\label{fig:labsam}
\end{figure}
\subsection{Optimization}
\label{sec:optim}
		\cgxxii{In this work, a label sampler~(illustrated in Fig~\ref{fig:labsam}) is proposed to introduce  \textoutofset{} samples and control complexity during training. 
		For each batch of training data, the sampler constructs a  \textcharset{} $\Ncharbatch\subset \Ncharspace$, composed  of  three disjoint subsets: the} ``positive'' characters  $\Ncharpos$, the ``negative'' characters $\Ncharneg$, and the special \cgxvii{tokens}\footnote{The end-of-speech label `[s]' and the unknown label `$[-]$'}. 
	
		Here, the module first collects all characters that appear in the batch, denoted as  $\Ncharlabel$. Then, \cgxv{it samples} a fraction of the $\Ncharlabel$ as the ``positive'' subset $\Ncharpos$  subjecting to two conditions. First, a fraction of characters shall be left unsampled as the out-of-set \cgxx{classes}, i.e.,    $|\Ncharpos|\leq|\Ncharlabel|*\Nfracseen$. Second, \cgxvii{the number of} all prototypes corresponding to $\Ncharpos$ should be smaller than the batch size limitation $\Nmaxtempbat$. 
	To cover more characters in the label space $\Ncharspace$, characters that don't  appear in the labels ($\Nctrain{}-\Ncharlabel$) are sampled into the ``negative'' subset $\Ncharneg$, with its size limited under $\Nmaxtempbat-|\Ncharpos|$ to keep the computation cost under control.  %
Finally, $\Ncharpos$,  $\Ncharneg$, and special labels are merged \cgxx{as} the final \textcharset{} $\Ncharbatch$ \cgxxi{for the iteration.}
The \cgxx{label} of each sample is then filtered \cgxxii{with $\Ncharbatch$}, by replacing all characters not in $\Ncharbatch$ with the \textoutofset label `$[-]$'.
The framework is trained with the images, filtered labels, and templates associated with $\Ncharbatch$. \cgxxii{Like most attention-based word-level text recognition methods~\cite{DAN,www,ASTER}, a cross-entropy loss $\Nclsloss$ is used} for the classification task, 
\begin{equation}
	\Nclsloss = -\frac{1}{\Ntimeupr}\sum_t^{\Ntimeupr}log(\sum_i^{|\Ncharspace|+2}\frac{Y_\indexof{\Ntime,i}e^{\Nfinscr_\indexof{l,i}}}{\sum_{j}^{|\Ncharspace|+2}e^{\Nfinscr_\indexof{\Ntime,j}}}),
\end{equation}	
where $Y_\indexof{\Ntime}\in\{0,1\}^{|\Ncharspace|+2}$ is the one-hot encoded label at timestamp $l$. \cgxxii{$\Ntimeupr$ indicates the number of characters in the sample (including the EOS character `[s]'), and the maximum length allowed by the model is denoted as $\Nmaxtime$.}

A regularization term $\Nembloss$ is adopted to improve the margins between classes,
\begin{equation} 
	\begin{split}
		\Nembloss=& \sum_{i,j}^{|P|}{Relu((P^{T}P)_\indexof{i,j}-m_p)}.
	\end{split}
\end{equation} 
Here $m_p$ is the cosine similarity between \cgxxi{the nearest pairs from} $n$ evenly distributed prototypes on a $\Nfeatdim$ dimension space. ``Evenly distributed'' means the distance between each prototype and its nearest neighbor equals the same number. Here, the margin, $m_p$, is used to preserve space for novel classes, and can be approximately solved via gradient descent. We estimate $m_p$ by optimizing the following formula,
\begin{equation} 
	\begin{split}
		Q^{*}&=\arg\min_{Q} \max_{(i,j)}(Q^{T}Q-2I)_{i,j}, \\
		m_p&= \max_{(i,j)}(Q^{*T}Q^{*}-2I)_{i,j},
	\end{split}
\end{equation} 
where $Q$ is a $\Nfeatdim \times n$ matrix with the random initialization, each column in $Q$ is a unit vector, $n$ is the estimated number of prototypes in the open set, $\Nfeatdim$ is the dimension of the prototypes, and $I$ is an identity matrix.  In this work, $\Nfeatdim$ is set to $\Sfeatdim$ and  N to $50,000$. That is to say, we estimate that character space $\Ncharspace$ contains $50,000$ distinct ``cases''. The optimization result of Eq. 14 suggests  $m_p$ is around $0.14$.   

The object function $\Nmodloss$ is the combination of the classification loss $\Nclsloss$ and the regularization term $\Nembloss$, and there is 
\begin{equation}	
	\Nmodloss=\Nmodloss+\Nwemb\Nembloss.
\end{equation}

\section{Experiments}
We conduct a variety of experiments to validate the proposed framework on three tasks. 
First, experiments on zero-shot Chinese character recognition provide a referenced comparison over the recognition capability on novel characters. 
Second, our framework is compared to conventional state-of-the-art close-set text recognition methods on standard benchmarks to validate its \cgxxii{feasibility} in conventional applications.
\cgxvii{Third}, \cgxix{a variety of} experiments on the open-set text recognition task are conducted and analyzed.
\cgxvii{In addition, extensive ablative studies are conducted to validate the effectiveness of each individual module proposed.}
Experiments show that our \mname{} framework can adapt to all three tasks without modification. 
\subsection{Implementation Details}
\label{sec:idet}
	Our implementation is based on the Decoupled Attention Network (DAN)~\cite{DAN}.
	Codes for our method and the dataset are also made publicly available on Github~\footnote{\url{https://github.com/lancercat/OSOCR}}.
	The regular model of our framework is mostly identical to DAN \cgxix{except that we remove the RNNs and add the additional modules.}  For the large model, we expand the \cgxx{channels of the feature extractor} to $1.5$ times the original \cgxxii{and  change the input to augmented colored images. 
	For hyperparameters,} 
	$\Nmaxtime$  is set \cgxx{to} $2$ for zero-shot Chinese character recognition (Sec.~\ref{zsocr}),  $25$ for close-set scene text recognition (Sec. ~\ref{csocr}), and $30$ for open-set scenarios (Sec. ~\ref{osocr}) because Chinese and Japanese scripts are generally longer.
	For the newly introduced hyperparameters, $\Nhardness$ of TPTNet is set to $\Shardness$ to reduce training variance by limiting the extent of spatial transformation.   
	\label{rev:sampler}
	 In the label sampler, the fraction of seen characters $\Nfracseen$ is set to $\Sfracseen$, and the maximum number of templates in an iteration $\Nmaxitr$ is set to $512$. 
	The $\Nwemb$ in loss function is set to $\Swemb$ for being a regularization term. \cgxxii{In addition, data augmentation approach from SRN~\cite{SRN} is adopted for training large models.}
	All experiments in this paper can be trained on Nvidia GPUs with \cgxix{8GB memory} and evaluated on a GPU with 2GB memory.
 \subsection{Zero-Shot Chinese Character Recognition}
\label{zsocr}
Zero-shot character recognition is a special case of open-set text recognition, where \cgxxii{$\Nctrain \cap \Ncktest = \emptyset$, $\Ncutest=\emptyset$,} and the lengths of all samples equal to one. 
	Here, we benchmark our framework on the HWDB~\cite{HWDB} dataset and the CTW~\cite{ctw} dataset to provide a referenced comparison over the capability to recognize \cgxxii{\textmyUKC{} characters} (Table~\ref{tab:perfch}), following the zero-shot Chinese character recognition community. 
	The HWDB dataset is a single-character hand-written dataset collected from $1020$ writers. We use the HWDB 1.0-1.2 for training and the ICDAR13 competition dataset for evaluation. The CTW dataset is a scene text recognition dataset with character-level annotations. This dataset is more difficult than HWDB due to noise, complex background patterns, and low contrast.
	In this work, Zhang et al.'s~\cite{hde} split scheme for training and evaluation labels is adopted to achieve fair comparisons.~\footnote{As they do not release their exact split, we spilt the datasets according to their paper and our exact splits can be referred to our released source codes.} For the HWDB dataset,  we first randomly split $1000$ labels  (characters) as the  evaluation \textcharset{} $\Ncktest$, and then randomly sample $500$, $1000$, $1500$, and $2000$ labels from the remaining as the training \textcharset{} $\Nctrain{}$.  On the CTW dataset, we first split $500$ characters for testing and then randomly sample $500$, $1000$, $1500$, and $2000$ characters for training. 
\begin{table*}[!t]
	\caption{
		Zero-shot character recognition accuracy on HWDB and CTW datasets. * indicates online-trajectory required for recognition.}
	\begin{center}
		\resizebox{\linewidth}{!}{
			\begin{tabular}{c|c|c|c|c|c|c|c|c|c}
				\hline
				& & \multicolumn{8}{c}{Accuracy (\%)} \\
				\cline{3-10}
				& &\multicolumn{4}{c|}{HWDB}&\multicolumn{4}{c}{CTW}\\
				\cline{3-10}
				Method&Venue&\multicolumn{4}{c|}{\# characters in training set}&\multicolumn{4}{c}{\# characters in training set} \\
				\cline{3-10}
				& &500& 1000&1500 &2000& 500& 1000&1500 &2000\\ 
				\hline
				CM*~\cite{cm19}&ICDAR'19&44.68&71.01&80.49&\textbf{86.73}&-&-&-&-\\
				\hline
				DenseRan~\cite{denseran}& ICFHR'18 &1.70&8.44&14.71&63.8&0.12&1.50&4.95&10.08\\
				\hline
				FewRan~\cite{fewran}& PRL'19&33.6&41.5&63.8&70.6&2.36&10.49&16.59&22.03\\
				\hline
				HCCR~\cite{hde}&PR'20&33.71&53.91&66.27&73.42&23.53&38.47&44.17&49.79\\
				\hline
				Ours&-&\textbf{47.92}&\textbf{74.02}&\textbf{81.11}&85.72&\textbf{28.03}&\textbf{49.00}&\textbf{58.37}&\textbf{64.03}\\
				\hline	
			\end{tabular}
		}
		\label{tab:perfch}
	\end{center}	
\end{table*}	

\begin{figure}[t]
	\centering
	\includegraphics[width=0.7\linewidth]{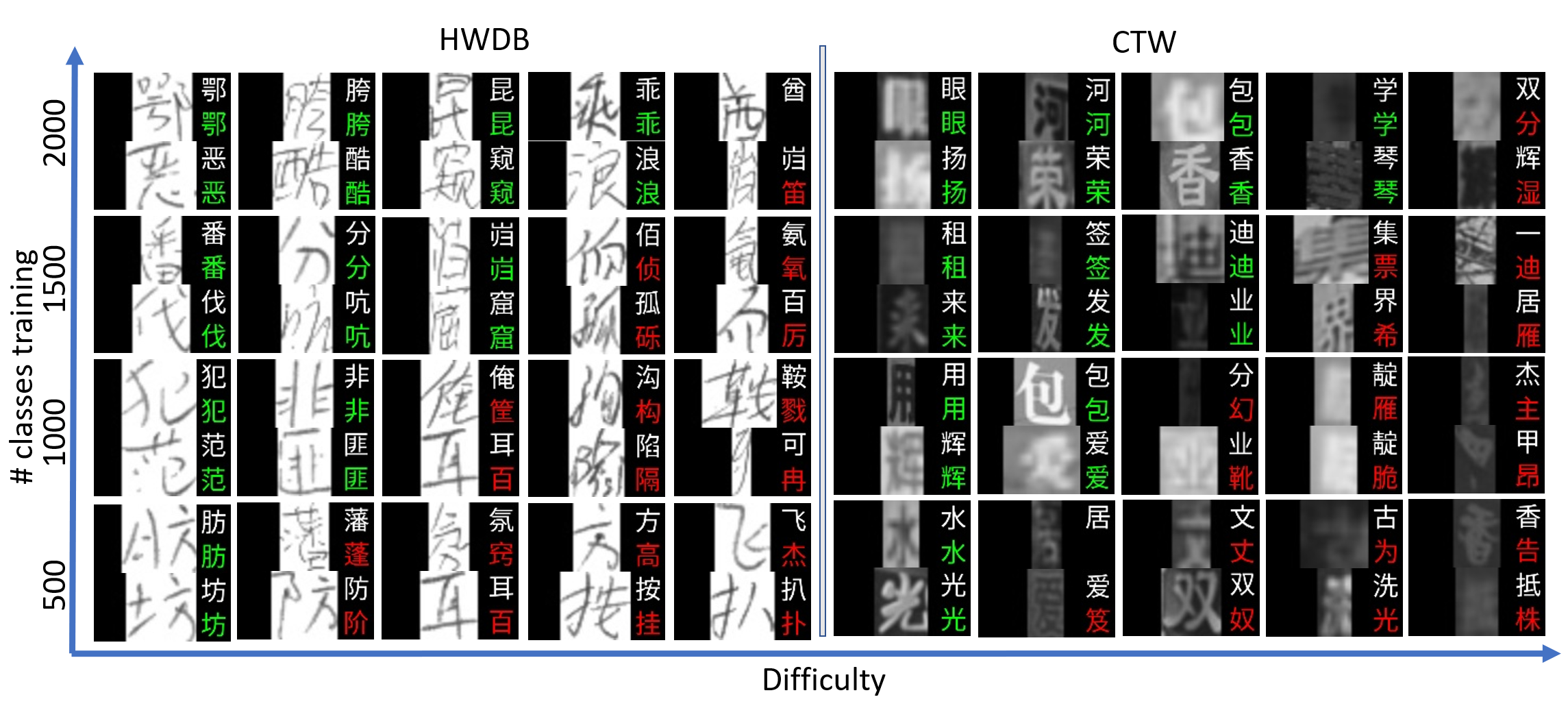}
	\caption{
		Zero-shot recognition results on the HWDB dataset and CTW dataset: Ground truth is annotated at the upper-right corner and prediction annotated at the bottom-\cgxviii{right} corner. Wrong predictions are indicated with the red color. Blank prediction indicates the model regards it as an unknown character.
	}
	\label{fig:dualperf}
\end{figure} %

\cgxxii{Quantitative performances are shown in Table~\ref{tab:perfch}, and qualitative results are shown in Fig.~\ref{fig:dualperf}.
	Results demonstrate decent robustness on novel characters for both hand-written and scene characters.  Quantitatively, the framework demonstrates more than $10\%$ accuracy advantages compared with a recent method~\cite{hde} for most setups on both datasets. Similar to other zero-shot methods, the robustness improves as the number of training characters increases, which means a more sufficient sampling on the label space $\Ncharspace$ improves performance.}
\cgxxii{Qualitatively, the model can handle a wide range of deterioration while showing some limitations. For handwritten data, the model may fail when a character is written in a cursive way or not correctly written (top-right sample of HWDB in Fig.~\ref{fig:dualperf}). For scene text data, the model has difficulty recognizing samples with severe blur and low-contrast.}
\begin{table}
	\caption{
		Open-set performance on HWDB and CTW datasets. \cgxxii{A stand for the Accuracy for \textmyUKC{} classes. R, P, and F stand for the Recall, Precision, and F-measure in spotting samples from \textmyUUC{} classes.}}
	\begin{center}\resizebox{0.65\linewidth}{!}{
			\begin{tabular}{c|c|c|c|c|c|c|c|c}
				\hline
				\cline{2-9}
				&\multicolumn{4}{c|}{HWDB}&\multicolumn{4}{c}{CTW}\\
				\hline
				\#\textmyUKC &100&200&400&500&50 &100&200&250\\
				\#\textmyUUC &900&800&600&500&450&400&300&250\\
				\hline
				A(\textmyUKC) &93.5&93.9&91.0&90.0 &79.3&77.1&72.6&69.6\\
				R(\textmyUUC) &48.0&24.6&7.9&5.1&73.3&54.7&37.7&31.5\\
				P(\textmyUUC) &99.7&99.5&97.9&96.7&98.9&95.9&92.4&88.6\\
				F(\textmyUUC) &64.8&39.5&14.6&9.7&84.2&69.7&53.5&46.5\\
				\hline
		\end{tabular}}
		\label{tab:coss}
	\end{center}
\end{table}
\cgxvii{We also extend these tasks to measure the open-set performance by further splitting different fractions of the evaluation characters into \cgxxii{$\Ncutest$ as \textmyUUC{}}.  The performances on the extended tasks are shown in Table~\ref{tab:coss}. Results show that our framework has some extent of rejection capability on both datasets.  On the other hand, the number of classes has a negative impact on the rejection capability and recognition accuracy}, which is also a common limitation in open-set recognition \cgxx{methods}~\cite{osr-survey}. 

Summarily, our proposed framework achieves significant advantages on all datasets and setups against most of the state-of-the-art methods, \cgxxii{and further provides some extent of rejection capability on ``out-of-set'' characters.  The framework also demonstrates competitive generalization ability against radical-based methods, justifying the feasibility to use glyphs as side information. Unlike radical-based frameworks, our method does not exploit any language-specific priors, thus not restricted to specific languages. As we only use the Noto-sans-regular font for glyphs, obtaining side-information for labels requires much fewer efforts compared to radical-based methods.}

\begin{table*}[!t]
	\caption{
		Performance on conventional close-set benchmarks, where * indicates character-level annotation required. 
	}
	\begin{center}
		\resizebox{\linewidth}{!}{
			\begin{tabular}{c|c|c|c|ccccc}
				\hline
				Methods&Venue&Training Set&RNN&IIIT5K&SVT&IC03&IC13&CUTE\\
				\hline
				AON~\cite{AON}&CVPR'18&MJ+ST&Y&87.0&82.8&91.5&-& 76.8\\
				ACE~\cite{ace}&CVPR'19&MJ&Y&82.3&82.6 &89.7&82.6 &82.6\\
				Comb.Best~\cite{www}&ICCV'19&MJ+ST&Y&87.9&87.5&94.4&92.3&71.8\\
				Moran~\cite{moran}&PR'19&MJ+ST&Y&91.2&88.3&95&92.4& 77.4\\
				SAR~\cite{SAR}&AAAI'19&MJ+ST&Y&91.5&84.5&-&-& 83.3\\
				ASTER~\cite{ASTER}&PAMI'19&MJ+ST&Y&93.4&\textbf{93.6}&94.5&91.8&79.5\\
				ESIR~\cite{ESIR}&CVPR'19&MJ+ST&Y&93.3& 90.2&-&-& 83.3\\
				SEED~\cite{seed}&CVPR'20&MJ+ST&Y&93.8&89.6&-&92.8& 83.6\\
				DAN~\cite{DAN}&AAAI'20&MJ+ST&Y&\textbf{94.3}&89.2&\textbf{95.0}&\textbf{93.9}& 84.4\\
				\hline
				Rosetta~\cite{Rosetta}\cite{www}&KDD'18&MJ+ST&N&84.3&84.7&92.9&89.0& 69.2\\
				CA-FCN*~\cite{cafcn}&AAAI'19&ST&N&92.0&82.1&-&91.4& 78.1\\	
				TextScanner*~\cite{ts}&AAAI'20&MJ+ST+Extra&N&93.9&90.1&-&92.9& 83.3\\

				\hline
				\textbf{Ours}&-&MJ+ST&N&90.40&83.92&91.00&90.24& 82.29\\
				\textbf{Ours-Large}&-&MJ+ST&N&92.63&88.25&93.42&93.79& \textbf{86.80}\\
				
				\hline
				
		\end{tabular}}
		\label{tab:perfcl}
	\end{center}
\end{table*}

\subsection{Standard Close-Set Text Recognition}
\label{csocr}
Experiments on popular close-set benchmarks are also conducted, and \cgxix{results} show that our method suffices as a feasible lightweight method on conventional close-set text recognition tasks.
\cgxvii{Following the majority \cgxix{of} methods in this community, we train our model on synthetic samples from Jaderberg (MJ)~\cite{MJ} and Gupta (ST)~\cite{ST}. We use}
the IIIT5K-Words (IIIT5K), Street View Text (SVT), ICDAR 2003 (IC03), ICDAR 2013 (IC13), and CUTE80 (CUTE) as our testing sets. 
Among the testing sets, IIIT5K, IC03, IC13, and SVT focus on regular shaped texts, and CUTE focuses on irregular-shaped text. IIIT5K contains $3,000$ testing images collected from the web. SVT has $647$ testing images from Google Street View. IC03 includes $867$ words from scene text, while IC13 extends IC03 and contains $1015$ images. CUTE80 includes $288$ curved samples. \cgxx{All models are} trained for $5$ epochs for close-set experiments\cgxvii{, and the results are shown in Table~\ref{tab:perfcl}.}

On the close-set benchmarks, our regular method is performance-wise better than early lightweight methods like Rosetta~\cite{Rosetta}. The method is also comparable to \cgxxi{the} state-of-the-art RNN-free method CA-FCN~\cite{cafcn}, while being significantly faster. Moreover, our method does not require character-\cgxxii{level} annotations for training. 
We also provide a larger model around the speed of CA-FCN~\cite{cafcn} yielding better performances close to heavy RNN-based methods~\cite{DAN}.
Plus, our methods is RNN-free, hence do not require batching up to margin out the latency caused by RNNs. 
\begin{table}[t]
	\begin{center}
		\caption{Experiments on lexicon-based close-set benchmarks. $^c$ indicates close-set methods and $^{+}$ indicates datasets other than MJ and ST are used.}
		\resizebox{0.8\linewidth}{!}{
			\begin{tabular}{c|c|c|c|c}
				\hline
				Method & Venue &IIIT5k (small/medium)&IC03 (full)&SVT (50)\\
				\hline
				AON$^{c}$~\cite{AON}& CVPR'18& 99.6/98.1&96.7&96\\
				ESIR$^{c}$~\cite{ESIR}& CVPR'19& 99.6/98.8&-&97.4\\
				CA-FCN$^{c+}$~\cite{cafcn}& AAAI'19& \textbf{99.8}/\textbf{98.9}&-&\textbf{98.5}\\
				\hline
				\hline
				Zhang et al.~\cite{eccv20}& ECCV'20 & 96.2/92.8&93.3&92.4\\
				\hline
				Ours&- &99.53/98.63&\textbf{96.77}&96.75\\
				\hline
		\end{tabular}}
		\label{tab:perfcld}
	\end{center}
\end{table}
\begin{figure}[h]
	\centering
	\includegraphics[width=0.7\linewidth]{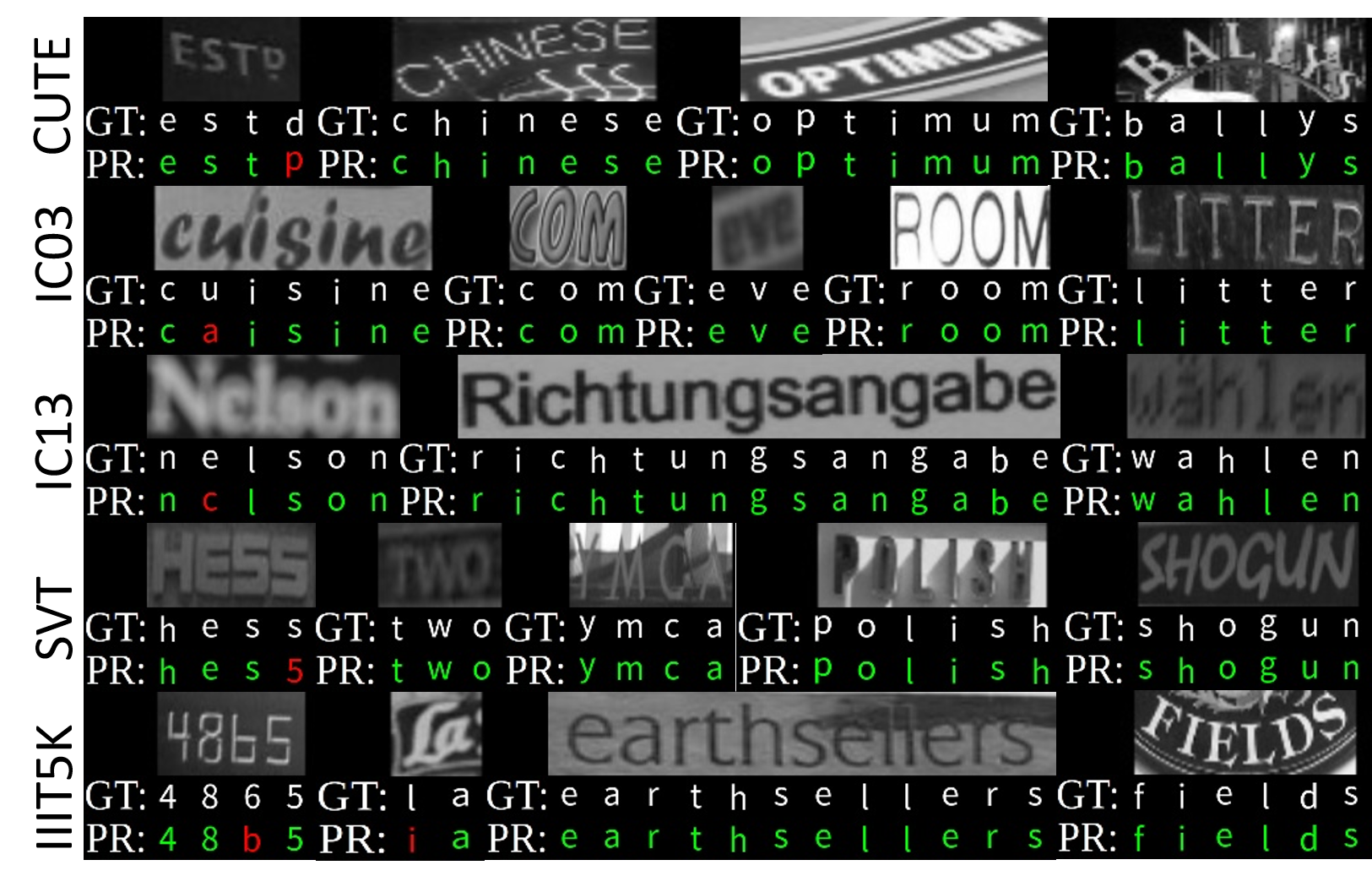}
	\caption{
		Recognition examples on the close-set benchmarks: Ground truth is annotated with ``GT'', prediction is annotated with ``PR'', and wrong predictions are indicated with the red color. 
	}
	\label{fig:perfcls}
\end{figure}

Representative samples are illustrated in Fig.~\ref{fig:perfcls}.  Our model demonstrates some extent of robustness for \cgxx{samples} with blur, irregular shapes, and different styles. 
\cgxxi{However, due to the absence of RNN modules, the framework lacks \cgxxii{sufficient} capability of modeling context information, hence showing a tendency} to confuse characters that have closer shapes, e.g., `s' and `5'. 
We also conduct a dictionary-based test to compare with other methods (Table~\ref{tab:perfcld}).  
\cgxvii{Results show \cgxix{a} significant performance advantage compared to Zhang et al.'s method~\cite{eccv20},  which is the only zero-shot text recognition method that tested on  standard close-set recognition benchmarks to our knowledge.  \cgxx{Under this setup, the framework is also performance-wise close to SOTA methods like CA-FCN~\cite{cafcn}.}
}

\begin{table}[!t]

	\begin{center}
		\caption{
			Speed evaluation on close-set benchmarks, where '*' means the results are derived from~\cite{www}.}
					\resizebox{\linewidth}{!}{\begin{tabular}{c|p{1cm}|c|c|c|c|p{1cm}|p{1cm}}
			\hline
			Method&Batch size&IIIT5K &CUTE&GPU&TFlops& Speed (ms)&Vram (MB)\\
			\hline
			CRNN*~\cite{CRNN}&1&78.2& ---&P40& 12&\textbf{4.4}&-\\
			\hline
			Rosetta*~\cite{Rosetta}&1&84.3& 69.2&P40& 12&4.7&-\\

			\hline
			Comb.Best* ~\cite{www}&1&87.9& 74.0&P40& 12&27.6&-\\
			\hline
			CA-FCN ~\cite{cafcn}&1&92.0& 79.9&Titan XP& 12&22.2&-\\
			\hline
			
			\hline
			Ours&1&90.40 &82.29&RTX2070M& 6.6&9.19&1227\\
			\hline
			Ours&16&90.40& 82.29&RTX2070M &6.6&4.68&1519\\
			\hline
			Ours-Large&1&\textbf{92.63}& \textbf{86.80}&RTX2070M& 6.6&14.18&1377\\
			\hline
			Ours-Large&16&\textbf{92.63}& \textbf{86.80}&RTX2070M& 6.6&7.81&1663\\
			\hline
			
		\end{tabular}}
		\label{tab:thru-lat}
	\end{center}
\end{table}
  \cgxx{The speeds of our framework with different  model sizes and batch sizes} are shown in Table~\ref{tab:thru-lat}. Speeds reported by other methods are also listed as a rough reference.  Our models run with FP32 datatype \cgxxi{on} an RTX2070 Mobile GPU, and the speeds are computed on the IIIT5k dataset. For maximum throughput, our method can reach a $213$ FPS running multi-batched. For latency critique tasks, our method can manage a $9.2$ ms latency under single-batched mode on \cgxiv{the GPU of around $7$ TFLOPS}. 
	Space-wise, our models do not require much \cgxx{VRAM} either. In summary, our model is reasonably small and fast, thus friendly to smaller devices like laptops, phones, and single-boards.

In conclusion, despite showing an acceptable margin against heavy \cgxx{SOTA methods, the framework} is comparable to, or better than \cgxxiii{many popular} RNN-free methods for close-set text recognition.  \cgxvii{Our framework demonstrates much more readiness to replace conventional text recognition methods in real-world applications, compared to other zero-shot text recognition methods like~\cite{eccv20}.}

\subsection{Open-Set Text Recognition}

\label{osocr}
\cgxxii{
In this section, we report the performance of the large model under five different splits, namely the GZSL mode, the OSR mode (w/o \textmyKUC{}), the OSR mode (with \textmyKUC{}), the GOSR mode, and the OSTR mode. The results and specifics of each split are shown in Table \ref{tab:oss-sets}. 
The model shows overall acceptable recognition capability for \cgxxiii{seen and novel} characters under most scenarios. For the rejection capability, the performance goes well with seen characters (\textmyKUC{}). The performance is also acceptable for some novel characters which are structural-wise close to characters in training set. However, the model demonstrates major limitations on the scalability over class number of $\Ncktest$. 
However, the precision of the framework is above 60\% on all setups, meaning human labor necessary to dismiss fake warnings is low. Considering most applications involves a large amount of data and each novel character only needs to be spotted once, the weakness on recall is not of much importance. Hence, the framework is still feasible for finding out novel  characters in the data-stream.
}

\begin{table*}[t]
	\caption{
		Result for different open-set character recognition splits.  \cgxxii{\NLineACR{} stands for the Line Accuracy for samples containing only $C^k_{eval}$ characters. R, P, and F stand for the Recall, Precision, and F-measure  performance spotting samples with out-of-set characters.}
	}
	\begin{center}\resizebox{\linewidth}{!}{
		\begin{tabular}{p{2cm}|p{3.5cm}|p{2cm}|c|c|c|c|c}
			\hline
			Name&$\Ncktest$ & $\Ncutest{}$& $|\Ncktest|$&\NLineACR{}&R&P&F\\
			\hline
			GZSL&{ Unique Kanji, Shared Kanji, Kana, Latin}&$\emptyset$&1460&30.83&-&-&-\\
			\hline
			\cgxxii{OSR w/o~\textmyKUC{}}& Shared Kanji, Latin &{Unique Kanji, Kana}&849&74.35&11.27&98.28&20.23\\	
			\hline
			\cgxxii{OSR with~\textmyKUC{}}& Shared Kanji &{Unique Kanji, Kana, Latin}&787&80.28&25.15&99.26&40.13\\
			\hline
			\cgxxii{GOSR}& Shared Kanji, Unique Kanji, Latin &{Kana}&1301&56.03&3.03&63.52&5.78\\	
			\hline
			\cgxxii{OSTR}& Shared Kanji, Unique Kanji &{Kana, Latin}&1239&58.57&24.46&93.78&38.80\\		
			\hline
			
		\end{tabular}
		\label{tab:oss-sets}
	}\end{center}
\end{table*}

\begin{figure}[!t]
	\centering
	\includegraphics[width=0.7\linewidth]{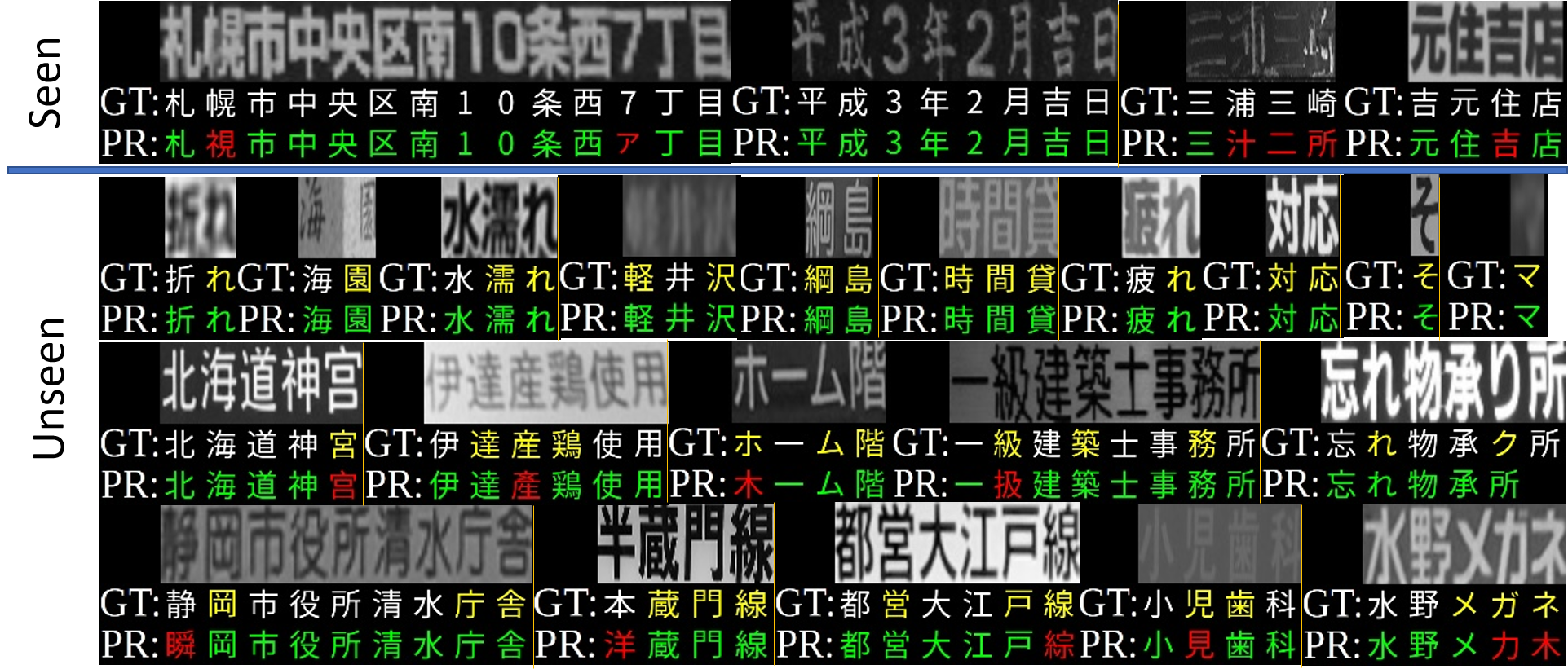}
	\caption{
		Good and failed cases in open-set text recognition experiments. Our model works better with shorter scripts and Kanji, while still bad in complex situations with multiple occurrences of Hiraganas and Katakanas.  Ground truth is annotated with ``GT'', prediction is annotated with ``PR'', and wrong predictions are indicated with the red color. Yellow characters in GT indicate that characters appear in the training set and white indicates seen characters. Correct predictions are indicated with the green color and prediction errors are highlighted with the red color.
	}
	\label{fig:goodNbad}
\end{figure}

\begin{table}[h]
	\caption{
		Result breakdown for open-set character recognition.
	}
	\begin{center}
		\resizebox{\linewidth}{!}{
		\begin{tabular}{c|c|c|c|c}
			\hline
			Name& Sample Requires & Sample Excludes &CA(\%)&\textbf{\NLineACR{}(\%)}\\
			\hline
			Shared Kanji&Shared Kanji& Unique Kanji, Kana&85.69&73.21\\
			\hline
			Unique Kanji&Unique Kanji&Kana &76.50&40.87\\	
			\hline
			All Kanji&Unique Kanji or Shared Kanji&Kana &79.94&54.91\\	
			\hline
			Kana&Hiragana or Katakana&&25.10&0.72\\
			\hline
			All&&&54.03&30.83\\	
			\hline	
		\end{tabular}}
		\label{tab:break}
	\end{center}
\end{table}

In this work, we primarily focus on implementing the language-agnostic recognition capability of unseen \cgxx{labels}, so we primarily analyze the detailed performance over the GZSL spilt.
\cgxxii{ Specifically,} qualitative samples are shown in Fig.~\ref{fig:goodNbad}, and a detailed breakdown of the large model is listed in Table~\ref{tab:break}. 
 In the table, ``Sample Requires'' indicates each sample must include at least one type \cgxx{among the listed}, while ``Sample Excludes'' \cgxxii{ refers to} each sample that cannot include any of the listed types of characters. ``Shared Kanji''  \cgxxii{refers to} Kanjis covered by the Tier-1 Simplified Chinese characters~\cite{HWDB}, while ``Unique Kanji'' \cgxxii{means unseen} Kanjis that are novel to the model. Despite showing reasonable robustness on the unique Kanjis, our framework shows limited generalization \cgxx{capability} on Hiragana and Katakana characters.
 
\begin{figure}[!t]
	\centering
	\includegraphics[width=0.7\linewidth]{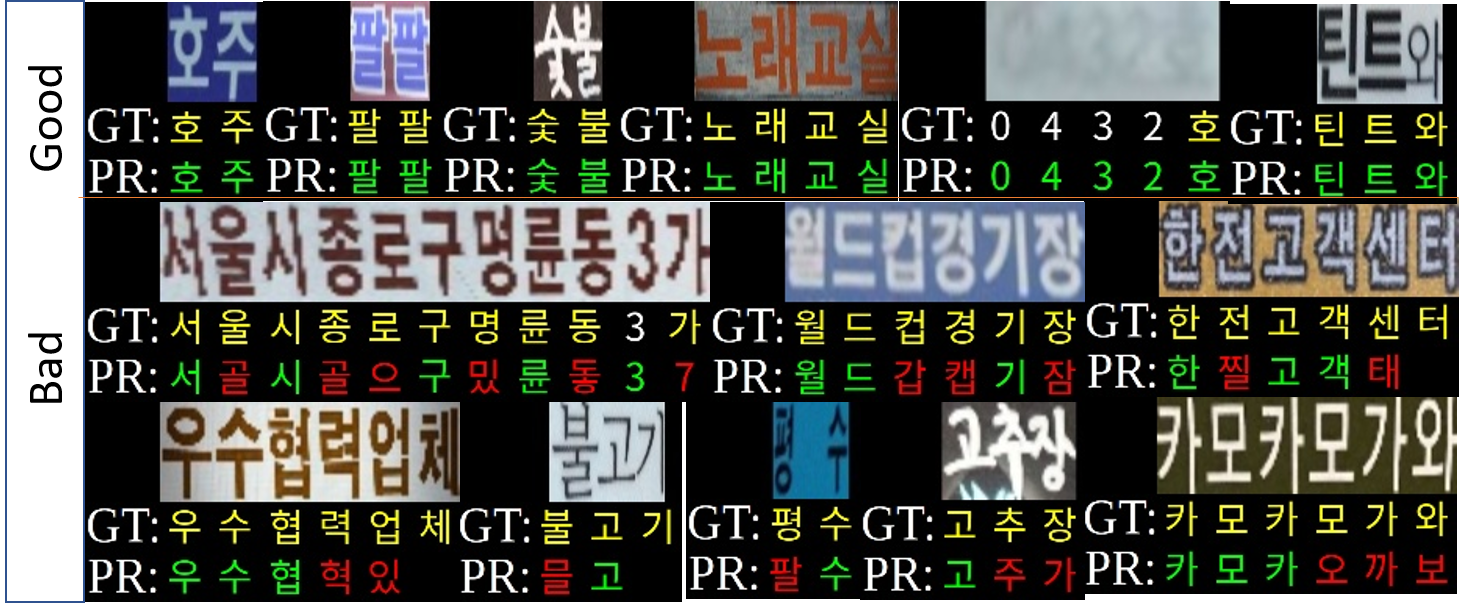}
	\caption{Qualitative results on the Korean language. Ground truth is annotated with ``GT'', prediction is annotated with ``PR''. Green indicates correct predictions red for wrong ones.}
	\label{fig:gangnam}
\end{figure}
\label{rev:oss_bench}

\begin{figure}[t]
	\centering
	\includegraphics[width=0.6\linewidth]{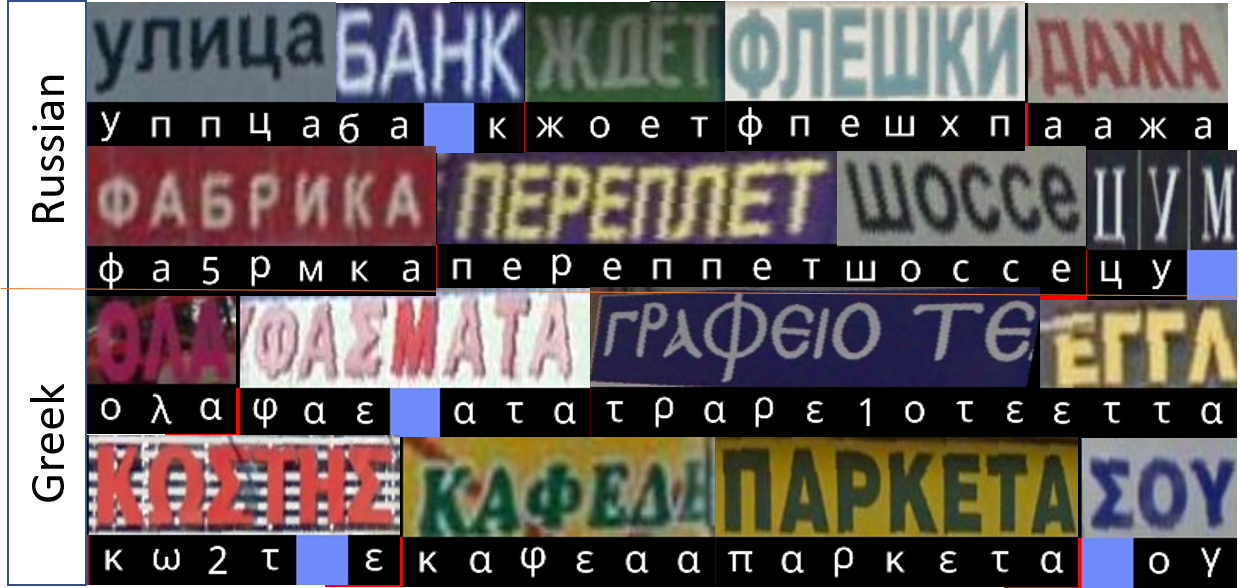}
	\caption{Qualitative results on the Russian scripts and Greek scripts from the SIW-13~\cite{SIW13} dataset. Recognition results are cast to lower case. Blue block means the character is rejected by the model.}
	\label{fig:rusputin-zero}
\end{figure}

\cgxviii{Note the size of \cgxix{the} character-set $|\Ncktest|$ can lay a serious impact on recognition performance. 
	\cgxxi{E.g., Row 2 and Row 3 in Table~\ref{tab:oss-sets} report higher \NLineACR{}s than Row 1 and Row 3 in Table~\ref{tab:break} for having a smaller $\Ncktest$.} This phenomenon suggests the decision boundaries of different prototypes may overlap with each other despite being bounded by the threshold $\Nradius$.  \cgxxii{Worth mentioning, that the result is not directly comparable} to the zero-shot character recognition results on the CTW dataset due to \cgxx{the vast differences among datasets, \textcharset{}s, and metrics.}}\label{rev:applenorange}

To further validate the generalization capability of the large model, we conducted evaluations on word-level samples from Korean, Russian, and Greek. The Korean words are drawn from the MLT dataset like the Japanese language, and qualitative results are shown in Fig.~\ref{fig:gangnam}. Quantitative-wise, the method shows $9.07$\% Character Accuracy and $1.35$\% Line Accuracy on $5,171$ Korean words, where the main limitation comes from confusion over characters with close shapes. For the Russian and the Greek language, we collected the samples from the SIW-13~\cite{SIW13} dataset. As the dataset does not provide annotation on contents, we only demonstrate the qualitative results in Fig~\ref{fig:rusputin-zero}.

\subsection{Ablative Studies}

To validate the robustness of the proposed modules, we performed extensive ablative studies on all three tasks, i.e., zero-shot Chinese character recognition, \cgxx{close-set text recognition,} and open-set text recognition. 
\begin{figure}[t]
	\centering
	\includegraphics[width=0.7\linewidth]{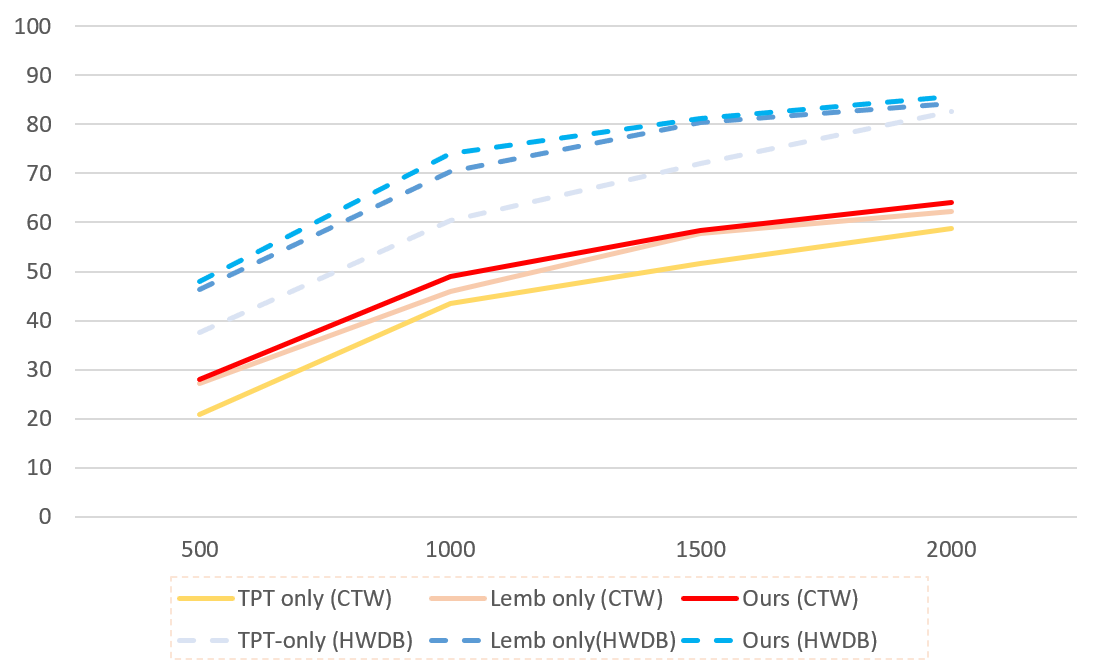}
	\caption{
		Ablative studies on $\Nembloss$  and TPTNet for zero-shot character recognition: X-axis indicates the number of training classes, and Y-axis indicates recognition accuracy.
	}
	\label{fig:abl-ctw}
\end{figure}
We first perform ablative studies on the zero-shot character recognition \cgxx{benchmarks} to validate the feasibility of $\Nembloss$ \cgxiv{on improving the generalization ability.} Parameters and set-ups are the same as the ones in the above experiments. \cgxx{Results} (Fig.~\ref{fig:abl-ctw}) show that $\Nembloss$ improves the performance significantly, \cgxx{especially with less training data.}
Our assumption is, that pushing prototypes apart makes the ProtoCNN more detail-sensitive, hence more robust to \cgxx{potential confusions} caused by characters with similar shapes. Additionally, the experiments also show that the topology-preserving transformation provides a limited improvement on character recognition tasks. \cgxiv{The reason could be precisely located characters are less prone to geometry deteriorations.}

\begin{table}[t]
	\begin{center}
		\caption{
			Ablative studies for the topology preserving transformation on conventional close-set text recognition benchmarks.
		}
		\begin{tabular}{c|c|c|c|c|c|c|c}
			\hline
			Method&TPT &$\Nembloss$&IIIT5K & SVT&IC03 &IC13&CUTE \\
			\hline
			Ours-Large&\checkmark&\checkmark&\textbf{92.63}  &\textbf{88.25}& \textbf{93.42} & \textbf{93.79}& \textbf{86.60}\\
			\hline
			Ours-Large w/o TPT&&\checkmark&91.80 & 86.08& 92.84  &92.21& 81.94\\
			\hline
		\end{tabular}
		\label{tab:ablcs}
	\end{center}
\end{table}
 \chgele{We then perform ablative studies to validate the effectiveness of the proposed topology-preserving transformation (TPT)  in scene text recognition. \cgxxi{Results show that the TPT module} yields a noticeable improvement on most datasets under 
the large configuration.
 Noticeably, TPT yields significant improvement on the CUTE dataset, which contains a lot of irregular-shaped text.}

\begin{table}[t]	
	\caption{Ablative study on the open-set text recognition dataset.}
	\begin{center}
		\begin{tabular}{c|c|c|c|c||c}
			\hline
			Method&TPT&$\Nembloss$&L2P& \NLineACR{} &\NLineACR{}-large \\
			\hline
			Ours&\checkmark&\checkmark&\checkmark&\textbf{28.11}&\textbf{30.83}\\
			Ours w/o TPT&&\checkmark&\checkmark&27.56&30.35\\
			Ours w/o $\Nembloss$&\checkmark &&\checkmark&24.89&29.23\\
			\hline
			Conventional&\checkmark&-&&17.63&18.05\\
			\hline
		\end{tabular}
		\label{tab:ablos}
		\label{tab:perfOST}
	\end{center}
\end{table}	 %

\cgxxii{
Finally, we perform ablative studies on the proposed open-set text recognition, with results shown in~\cgxxii{Table~\ref{tab:perfOST}}. 
Module-wise, both the proposed topology-preserving transformation (TPT) and the \cgxxiii{prototype} regularization \cgxxi{term} ($\Nembloss$) improve the recognition performance on the open-set challenge.  Framework-wise, the full framework yields significantly better performance against the conventional method over both character accuracy and line accuracy. Here, the conventional method replaces the label-to-prototype module and the open-set predictor with a linear classifier, pipeline-wise similar to an RNN-free version of DAN~\cite{DAN}. 
This indicates our framework achieves reasonable recognition capability over novel characters without sacrificing much close-set performance.
}

In summary, our label-to-prototype learning framework achieves competitive performance on novel characters without sacrificing much of the close-set performances. The topology-preserving transformation module shows robust improvement on all setups in all three tasks. 
The regularization term $\Nembloss$ also shows steady improvements on novel characters,  potentially by improving the inter-class distance on the prototype space, leaving more space for novel classes.

\begin{figure}[t]
	\centering
	\includegraphics[width=0.5\linewidth]{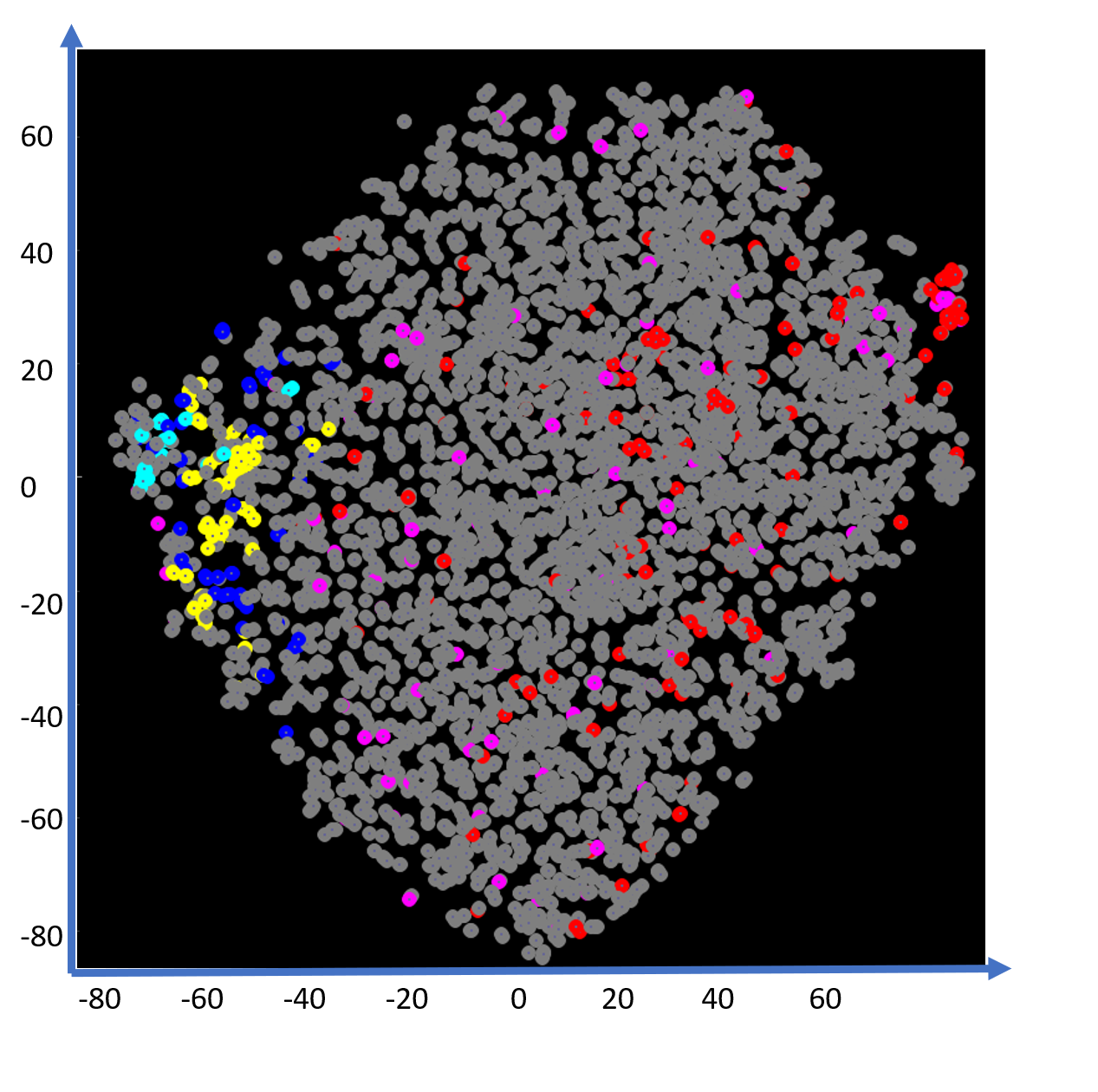}
	\caption{
		The t-SNE visualization on ProtoCNN outputs: Gray indicates $3755$ simplified Chinese characters and English letters, and red indicates other valid Chinese characters. Purple indicates unique characters in Kanji, blue and yellow for Hiragana and Katakana, and cyan for symbols.
	}
	\label{fig:tsne}
\end{figure}

\section{Limitation}

The major problem of our framework is the limited robustness over domain bias, e.g, the performance drop on Hiragana and Katakana.
Possibly due to the significant visual differences between Kanas and characters in the training set. \cgxxii{Specifically, Kanas} reside in an under-sampled subspace on the template space  $\Ntempspace$, which \cgxxii{consequentially} leads to larger ``errors'' in generated  prototypes.
This assumption can also be backed by the t-SNE~\cite{tsne} visualization of generated prototypes (shown in Fig.~\ref{fig:tsne}).
We can see that \cgxxi{Kanas} gather in the left-most region of the oval spanned by the training set characters (gray dots), while most other novel characters like Unique Kanjis are distributed relatively evenly in the space. %
Since the network training is a function fitting process, the points from less-sufficiently sampled regions tend to have larger fitting errors. In this case, the under-sampling problem causes \cgxxi{bad class centers and visual features, resulting in lower recognition accuracy of Kanas and Korean Hanguls.}

\section{Conclusions}
We propose a novel \cgxx{framework} for the open-set text recognition task via label-to-prototype learning. \chgtena{Results show} a largely improved performance compared with state-of-the-art methods on zero-shot character recognition tasks. Our framework also reaches competitive performance on the conventional close-set text \chgtena{line} recognition benchmarks, demonstrating its readiness for applications. Moreover, our method draws a strong baseline on the open-set text recognition task with a large \textcharset{} and complex deterioration. %
\chgtena{ In summary, our method can robustly handle isolated characters and text lines in both open-set and close-set scenarios.}

Despite showing impressive generalization capability on open-set scenarios, our \cgxxi{framework} still has some limitations. Specifically, our method tends to fail to recognize characters  that have significantly different shapes from the seen characters \cgxx{(e.g., Kanas and Hanguls)}. 
Domain \cgxxiii{adaption} techniques may be investigated to solve the domain bias among complex characters and different languages in future \cgxxi{research}. 
\cgxiii{Also, despite showing some extent of rejection capability, the model still has much room to improve.} \cgxx{This makes another topic of our future research.} 

\section{Acknowledgement}
	The research is partly supported by the National Key Research and Development Program of China (2020AAA09701), The National Science Fund for Distinguished Young Scholars (62125601), and the National Natural Science Foundation of China (62006018, 62076024).

{\small
\bibliographystyle{ieee_fullname}
\bibliography{cvpr.bib}
}
\clearpage
\begin{table}[t]	
	\caption{ Study on various factors affecting character recognition performance on  the CTW dataset.}
	\begin{center}
		\begin{tabular}{c|c|p{1.4cm}|p{1.4cm}|c|c|c|c|c}
			\hline
			Name&Distance&Training Template&Evaluation Template&$\Ncktest$& A&R&P&F \\
			\hline
			Ours&DotProd&Sans&Sans&500&65.53&-&-&-\\
			Ours-X&DotProd&Sans&Serif&500&53.52&-&-&-\\
			MF-A&DotProd&Sans+Serif&Sans&500&66.79&-&-&-\\
			MF-B&DotProd&Sans+Serif&Serif&500&66.72&-&-&-\\
			Cos&Cosine&Sans&Sans&500&59.53&-&-&-\\
			\hline
			Ours&DotProd&Sans&Sans&250&70.62&33.34&88.00&48.36\\
			Ours-X&DotProd&Sans&Serif&250&59.50&42.01&76.98&54.35\\
			MF-A&DotProd&Sans+Serif&Sans&250&69.89&30.94&85.63&45.46\\
			MF-B&DotProd&Sans+Serif&Serif&250&68.03&28.64&84.17&42.73\\
			Cos&Cosine&Sans&Sans&250&65.66&11.32&85.21&20.00\\
			\hline

		\end{tabular}
		\label{tab:osana}
	\end{center}
\end{table}	 
\section*{Appendix A}
In this section, we discuss  other factors that affect recognition  performance, namely the fonts used for templates, the metric used, and the number of characters in $\Ncktest$. Note the performance is a little bit different as the results are from another run on a faster device. 
For font selection for templates, we train the model with both Noto-sans and Noto-serif and test with each font (MF-A and MF-B). Extra fonts does not yield noticeable performance change, presumably due to the increased training variance. ``Ours-X'' evaluates the ``Ours'' model trained with Sans font with Serif font, yielding a significant performance degrade due to the shift on class centers. The rejection rate also increased due to the center shift, resulting in a higher recall and lower precision.  For distance metric, we implement the scaled cosine metric according to  Parisot et al.~\cite{scos}, however, the performance is not ideal. We will attempt to figure out the reason in the future. As we can see in Table~\ref{tab:coss}, the scalability of this method is limited like many classification methods, which is a common issue in the classification domain.

\section*{Appendix B}
\begin{figure}
	\centering
	\includegraphics[width=0.7\linewidth]{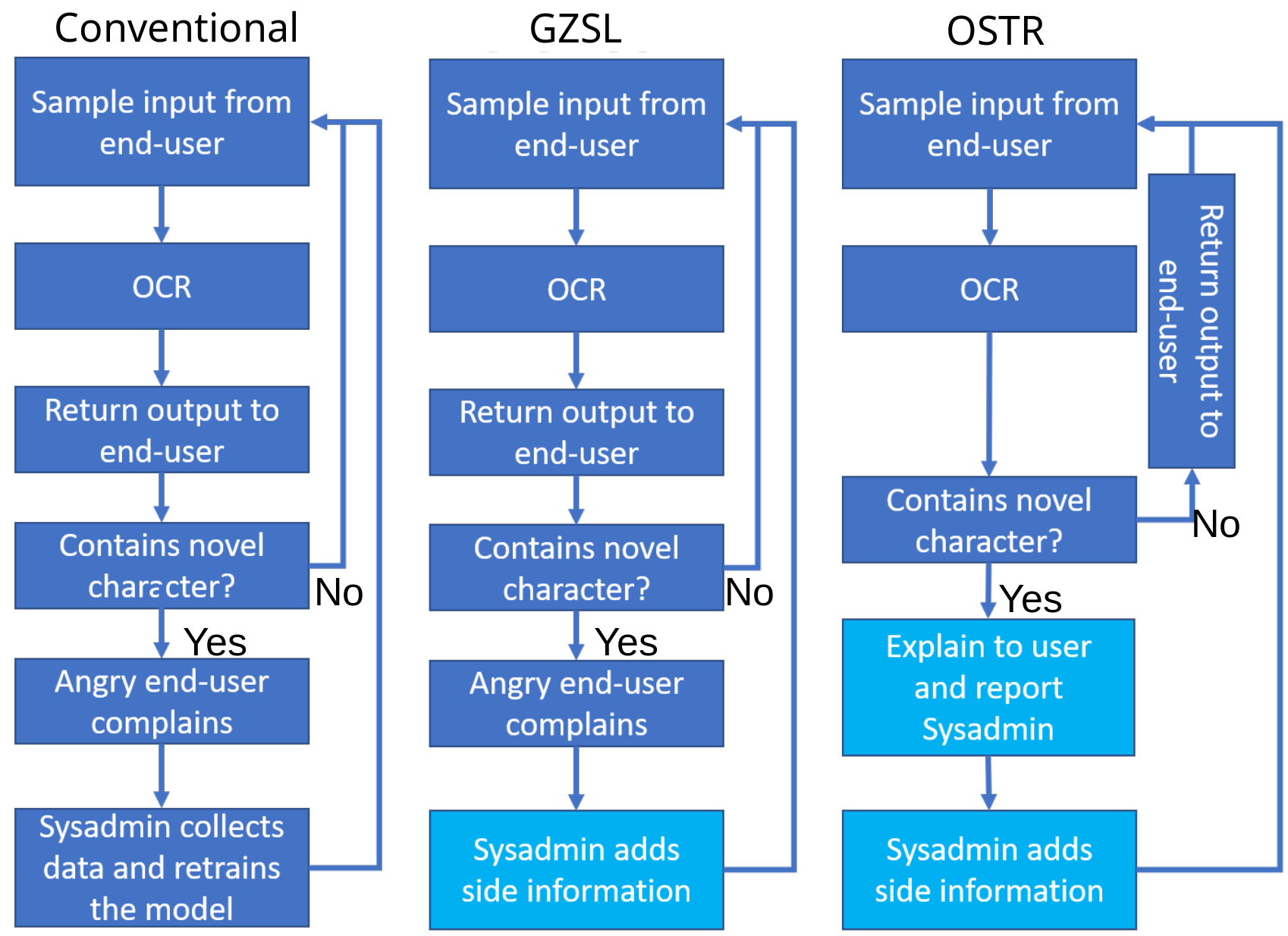}
	\caption{Comparison between the usage of close-set text recognition, zero-shot text recognition and \tname{} in applications.}
	\label{fig:altertask}
\end{figure}
In this section, we explain the reasons behind the definition of \tname{}, and corresponding use cases in real-life applications show in Figure~\ref{fig:altertask}. 
With the openness and fast-evolving nature of the internet, the character set of the incoming dataset can change over time. The volatile property of the character set also applies to historical document recognition as previously unknown characters could be found from the samples. In a word, the two main goals of the OSTR task are reducing adaption cost and achieving active novel characters spotting from the data stream.

Considering the majority of samples in real-life applications still involve seen characters, the task demands recognition capability of  \textmyKKC{} like conventional close-set text recognition~\cite{www}.
 However, close-set methods require a data collection (synthetic or annotation) and retraining process on each character set change.
Generally speaking, the retraining process would cost around several GPU days, depending on the amount of data and model size.  For the internet-oriented data, the cost could be hefty as the changes may occur frequently, under the pretext of the booming trend of mixing emojis with normal characters, and emojis happen to evolve constantly.  On the other hand, for historical data where annotation could be expensive and synthetic data can be limited due to insufficient amount of available fonts, rendering the conventional pipeline less feasible.
To reduce the adaption cost, we introduce the recognition capability of  \textmyUKC{} like zero-shot tasks~\cite{fewran,eccv20}.  

Furthermore, in real applications we do not know what novel characters we may encounter, nor do we know when. Hence, as a result of lacking rejection capability, the conventional close-set and zero-shot text recognition methods would need to wait till end-users~(OCR service subscribers) complain before the sysadmins~(OCR service providers) can notice the novel characters. 
To address this problem, we introduce the rejection capability over \textmyUUC{} characters like OSR tasks~\cite{osr-survey}.
 Here we further extend the subjects to seen classes (\textmyKUC{}) as well. One main reason is, that some characters or emoticons may get temporarily obsolete and become less frequently used.  In this case, the user may want to remove them from the character set to speed up, while not risking missing and mistaking them for something else when they re-appear. Hence, we naturally extend the subject of rejection to include both \textmyKUC{} and \textmyUUC{}.

Summing up,  the OSTR task demands the capability to match samples with side-information of an arbitrary set of classes for recognition, and reject samples failing matching any classes in the set.  Noteworthy, the task still demands close-set recognition performance, hence the sysadmins still retain the option to retrain an OSTR method with collected data. Furthermore, an OSTR method offers extra options of using a quick patch as a temporary~(till retraining is done) or permanent~(no retraining at all) solution.

\clearpage
\section*{Appendix C}
As this paper involves a lot of notions, we collect scalars in Table ~\ref{tab:notation_numbers}, sets in Table ~\ref{tab:notation_sets}, and other notations in Table ~\ref{tab:notation_other}. Each table illustrates the notations with their shapes, first appearances, and corresponding short descriptions. 
\begin{table*}
	\caption{Notations of ``numbers'' in the paper.}
	\begin{tabular}{c|c|c|P{2cm}|p{6cm}}
		\hline
		Notation& Occurrence &Type& Shape / Value &Description\\
		\hline
		$\Nfeatdim$&\Afeatdim&\Tfeatdim&$\Sfeatdim$&\Dfeatdim\\
		\hline
		$\Nhardness$&\Ahardness&\Thardness&$\Shardness $&\Dhardness\\
		\hline
		$\Nfracseen$&\Afracseen&\Tfracseen&$\Sfracseen $&\Dfracseen\\
		\hline
		$\Nmaxtempbat$&\Amaxtempbat&\Tmaxtempbat&$\Smaxtempbat $&\Dmaxtempbat\\
		\hline
		$\Nmaxtime$&\Amaxtime&\Tmaxtime&$ \Smaxtime$&\Dmaxtime\\
		\hline
		$\Nwemb$&\Awemb&\Twemb&$\Swemb $&\Dwemb\\
		\hline
		$\Nmaxitr $&\Amaxitr &\Tmaxitr &$\Smaxitr  $&\Dmaxitr \\
		\hline
		$\Ntime$&\Atime&\Ttime&$\Stime $&\Dtime\\
		\hline
		$\Ntimeupr$&\Atimeupr&\Ttimeupr&$ \Stimeupr$&\Dtimeupr\\
		\hline
		$\Nradius$&\Aradius&\Tradius&$ \Sradius$&\Dradius\\
		\hline
		$\Nembloss$&\Aembloss&\Tembloss&$\Sembloss $&\Dembloss\\
		\hline
		$\Nmodloss$&\Amodloss&\Tmodloss&$\Smodloss$&\Dmodloss\\
		\hline
		$\Nclsloss$&\Aclsloss&\Tclsloss&$\Sclsloss $&\Dclsloss\\
		\hline
		$\NLineACR$&\ALineACR&\TLineACR&$ \SLineACR$&\DLineACR\\
		\hline
		$\NCharACR$&\ACharACR&\TCharACR&$ \SCharACR$&\DCharACR\\
		\hline
		$\Nrec$&\Arec&\Trec&$ \Srec$&\Drec\\
		\hline
		$\Npre$&\Apre&\Tpre&$ \Spre$&\Dpre\\
		\hline
		$\Nfm$&\Afm&\Tfm&$ \Sfm$&\Dfm\\
			\hline
		$\Nanychar{i}$&\Aanychar{i}&\Tanychar{i}&$ \Sanychar{i}$&\Danychar{i}\\
		\hline
		
	\end{tabular}
	\label{tab:notation_numbers}
\end{table*}
\begin{table*}
	\caption{Notations of ``sets'' in the paper.}
	\begin{tabular}{c|c|c|P{2cm}|p{6cm}}
		\hline
		Notation& Occurrence &Type& Element Shape &Description\\
		\hline
		$\Ncharspace$&\Acharspace&\Tcharspace&$ \Scharspace$&\Dcharspace\\
		\hline
		$\Ntempspace $&\Atempspace &\Ttempspace &$\Stempspace  $&\Dtempspace \\
		\hline
		$ \Nprotospace$& \Aprotospace& \Tprotospace&$ \Sprotospace $& \Dprotospace\\
		\hline
		\hline
		$\Ngt$&\Agt&\Tgt&$\Sgt$&\Dgt\\
		\hline
		$\Npr$&\Apr&\Tpr&$\Spr$&\Dpr\\
		\hline
		$\Nctrain$&\Actrain&\Tctrain&$\Sctrain$&\Dctrain\\
		\hline
		$\Nctest$&\Actest&\Tctest&$\Sctest$ &\Dctest\\
		\hline
		$\Ncktest$&\Acktest&\Tcktest&$\Scktest$&\Dcktest\\
		\hline
		$\Ncutest$&\Acutest&\Tcutest&$ \Scutest$&\Dcutest\\
		\hline
		$\Ncharlabel$&\Acharlabel&\Tcharlabel&$\Scharlabel $&\Dcharlabel\\
		\hline
		$\Ncharbatch$&\Acharbatch&\Tcharbatch&$\Scharbatch $&\Dcharbatch\\
		\hline
		$\Ncharpos$&\Acharpos&\Tcharpos&$\Scharpos $&\Dcharpos\\
		\hline
		$\Ncharneg$&\Acharneg&\Tcharneg&$\Scharneg $&\Dcharneg\\
		\hline
		$\Nchar$&\Achar&\Tchar&$\Schar $&\Dchar\\
		\hline
		$\Nanytemps{i}$&\Aanytemps{i}&\Tanytemps{i}&$\Sanytemps{i} $&\Danytemps{i}\\
		\hline
	\end{tabular}
	\label{tab:notation_sets}
\end{table*}

\begin{table*}
	\caption{Other important notations in the paper.}
	\begin{tabular}{c|c|c|P{2cm}|p{6cm}}
		\hline
		Notation& Occurrence &Type& ``Shape'' &Description\\
		\hline
		$\Nsamefn$&\Asamefn&\Tsamefn&$\Ssamefn $&\Dsamefn\\
		\hline
		$\Nfned$&\Afned&\Tfned&$\Sfned $&\Dfned\\
		\hline
		$\Nfnlen$&\Afnlen&\Tfnlen&$\Sfnlen $&\Dfnlen\\
		\hline
		$\Nfnrej$&\Afnrej&\Tfnrej&$\Sfnrej $&\Dfnrej\\
		\hline
		$\NlTproto$&\AlTproto&\TlTproto&$\SlTproto$&\DlTproto\\
		\hline
		$\NlTtemp $&\AlTtemp &\TlTtemp &$\SlTtemp  $&\DlTtemp \\
		\hline
		$\NtTproto$&\AtTproto&\TtTproto&$\StTproto $&\DtTproto\\
		\hline
		$\Nproto{}$&\Aproto{}&\Tproto&$\Sproto $&\Dproto\\
		\hline
		$\Nfeatmap{.}$&\Afeatmap{.}&\Tfeatmap{.}&$\Sfeatmap{.} $&\Dfeatmap{.}\\
		\hline
		$\Ndensx{}$&\Adensx{}&\Tdensx&$\Sdensx $&\Ddensx\\
		\hline
		$\Ndensy{}$&\Adensy{}&\Tdensy&$\Sdensy $&\Ddensy\\
		\hline
		$\Nnewcord{}$&\Anewcord{}&\Tnewcord&$\Snewcord $&\Dnewcord\\
		\hline
		$\Nfeatseq$&\Afeatseq&\Tfeatseq&$ \Sfeatseq$&\Dfeatseq\\
		\hline
		$\Nanytemp{i}$&\Aanytemp{i}&\Tanytemp{i}&$\Sanytemp{i} $&\Danytemp{i}\\
		\hline
		$\Nanyproto{i}$&\Aanyproto{i}&\Tanyproto{i}&$\Sanyproto{i} $&\Danyproto{i}\\
		\hline
		$\Ncasesim$&\Acasesim&\Tcasesim&$\Scasesim $&\Dcasesim\\
		\hline
		$\Nchasim$&\Achasim&\Tchasim&$\Schasim $&\Dchasim\\
		\hline
		$\Nfinscr$&\Afinscr&\Tfinscr&$\Sfinscr $&\Dfinscr\\
		\hline
	\end{tabular}
	\label{tab:notation_other}
\end{table*}
\end{document}